\documentclass[sigconf]{acmart}
\usepackage{graphicx}

\usepackage{ifthen}
\newboolean{techreport}
\usepackage{caption}
\usepackage{multirow}
\usepackage{subcaption}
\usepackage{hyperref}
\usepackage{amsmath}
\usepackage{tikz}
\usepackage[]{footmisc}
\usepackage{graphicx}
\usepackage{pifont}
\usepackage{gensymb}
\usepackage{algpseudocode}

\usepackage[ruled,linesnumbered,vlined]{algorithm2e}
\usepackage{amsmath,amstext}
\let\oldnl\nl
\newcommand{\nonl}{\renewcommand{\nl}{\let\nl\oldnl}}
\usepackage{tikz}

\definecolor{codegreen}{rgb}{0,0.6,0}
\definecolor{codegray}{rgb}{0.5,0.5,0.5}
\definecolor{codepurple}{HTML}{C42043}
\definecolor{backcolour}{rgb}{1,1,1}

\newcommand\numberstyle[1]{%
    \footnotesize
    \color{codegray}%
    \ttfamily
    \ifnum#1<10 0\fi#1 |%
}

\newtheorem{definition}{Definition}[section]

\usepackage{enumitem}

\renewcommand\footnotetextcopyrightpermission[1]{} 
\newtheorem{problem}{Problem Definition}

\begin{document}
\setboolean{techreport}{true}

\title{SwarMer: A Decentralized Localization Framework for Flying Light Specks}

\author{Hamed Alimohammadzadeh, Shahram Ghandeharizadeh}
\affiliation{%
  \institution{University of Southern California}
\streetaddress{Computer Science Department}
 \city{Los Angeles}
 \state{California}
 \country{USA}
}
\email{{halimoha,shahram}@usc.edu}

\begin{abstract}
Swarm-Merging, SwarMer, is a decentralized framework to localize Flying Light Specks (FLSs) to render 2D and 3D shapes.
An FLS is a miniature sized drone equipped with one or more light sources to generate different colors and textures with adjustable brightness.  It is battery powered, network enabled with storage and processing capability to implement a decentralized algorithm such as SwarMer.
An FLS is unable to render a shape by itself.
SwarMer uses the inter-FLS relationship effect of its organizational framework to compensate for the simplicity of each individual FLS, enabling a swarm of cooperating FLSs to render complex shapes.
SwarMer is resilient to both FLSs failing and FLSs leaving to charge their battery.
It is fast, highly accurate, and scales to remain effective when a shape consists of a large number of FLSs.

Source code available at \url{https://github.com/flyinglightspeck/SwarMer}.  See \url{https://youtu.be/BIiBxD_aUz8} for a MATLAB demonstration of SwarMer,
\url{https://youtu.be/Lh11tWWOP5Y} for two relative localization techniques as SwarMer plugins.
SwarMer is able to transition FLSs from illuminating one point cloud to the next point cloud, see \url{https://youtu.be/4GhhlSq4UrM}.
It may start with a random pattern and illuminate a shape, see \url{https://youtu.be/cZrz0e61txU}.
The reported implementation of SwarMer localizes FLSs into a Hat (\url{https://youtu.be/y_xHdz5bs5M}), a Dragon (\url{https://youtu.be/PRyjdmw8NVc}), and a Skateboard (\url{https://youtu.be/a0fFu0z6BU0}).

\end{abstract}

\maketitle
\pagestyle{plain} 

\section{Introduction}

An FLS is a miniature sized drone equipped with one or more light sources to generate different colors and textures with adjustable brightness.  Synchronized swarms of FLSs will illuminate objects in a pre-specified 3D volume, an FLS display~\cite{shahram2021,shahram2022}.  The display may be a cuboid that sits on a table~\cite{dv2023} or hangs on a wall, the dashboard of a self-driving vehicle, a room, etc.  It will render static 2D or 3D point clouds by assigning the coordinates of their points to different FLSs to illuminate.

An FLS display has diverse multimedia applications ranging from entertainment to healthcare.
With entertainment, a photographer may capture 3D point clouds using a  depth camera, e.g., Intel RealSense D455, and view them using an FLS display.
With healthcare, a physician may illuminate and examine 3D MRI scans of a patient in real time using an FLS display.


Localizing FLSs to render a 2D or a 3D point cloud is non-trivial for several reasons.  First, in an indoor setting, FLSs may lack line of sight with GPS satellites to position themselves.
Even if GPS was available, it would be too coarse~\cite{swarmboids2015,leong2016,shahram2021}.  The best some GPS devices claim is a 3 meter accuracy 95\% of the time~\cite{leong2016,sun2021}.  

\begin{figure}
\centering
\includegraphics[width=3.4in]{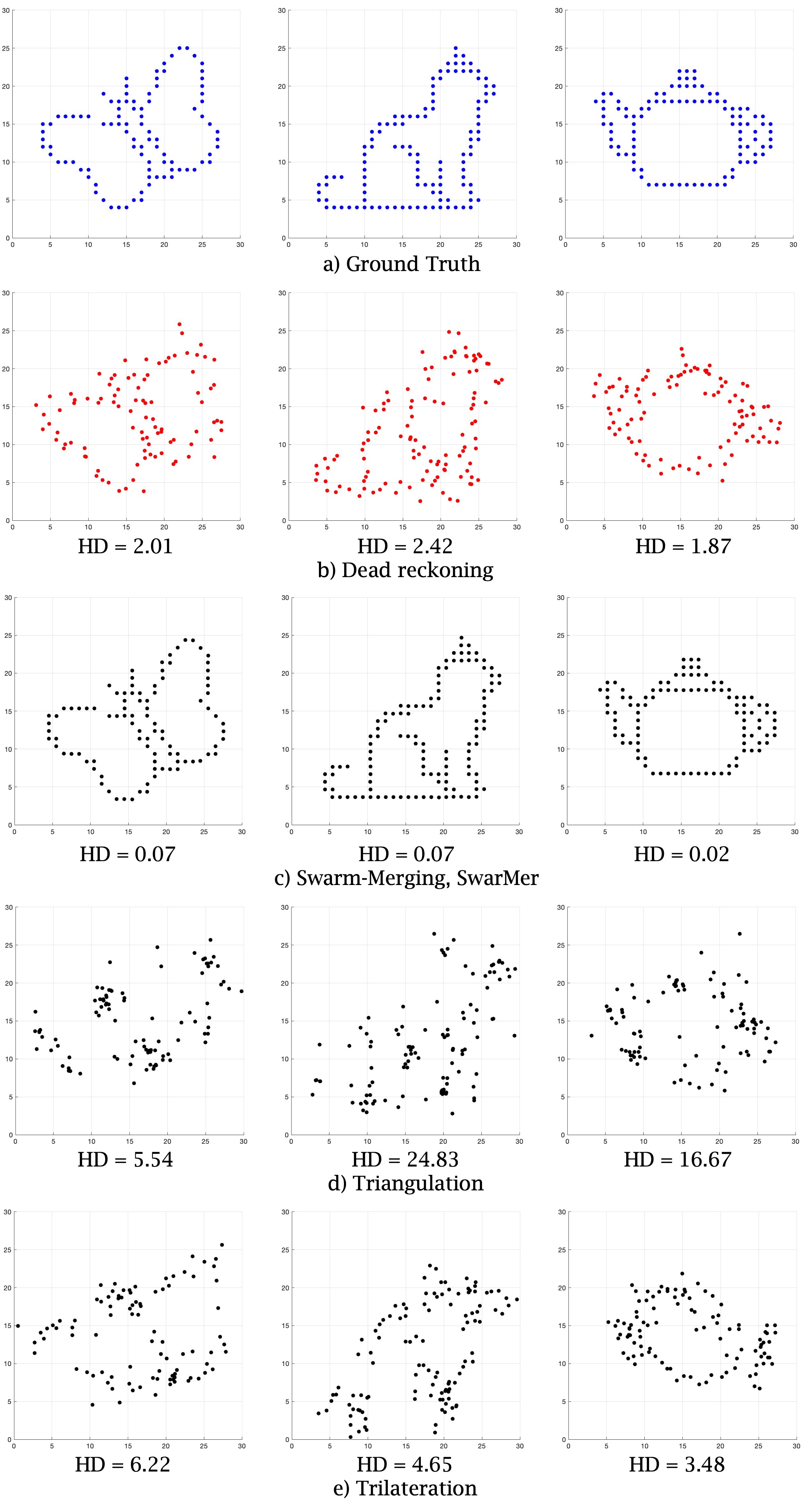}\hfill
\caption{(a) A 2D butterfly, cat, and teapot,
(b) illuminations with dead reckoning using $\epsilon$=5 degrees of error, (c) adjustments with swarm-merging using dead reckoning with $\epsilon$=5 degrees of error, (d-e) adjustment with Triangulation and Trilateration using 100\% accurate distance and angle.}
\label{fig:cmp}
\end{figure}

Second, a dead-reckoning technique that requires an FLS to fly to its assigned position is known to suffer from errors that increase as a function of the distance traveled by a drone~\cite{campos2021,shurin22,chenSurvey2022,deter2022}.  Greater distances result in higher errors, distorting the final rendering.
To illustrate, Figure~\ref{fig:cmp}.b shows rendering of three 2D point clouds of Figure~\ref{fig:cmp}.a (ground truth) using dead reckoning with 5 degrees of error.
(See Section~\ref{sec:dr} for a formal definition of the {\em degree of error}.)
The FLSs are deployed by a dispatcher at the origin, (0, 0).  Each FLS uses a collision free path~\cite{clearpath2009,collisionfree2012,collisionfree2015,collisionavoidance2018,ReactiveCollisionAvoidance2008,ReactiveCollisionAvoidance2011,ReactiveCollisionAvoidance20112,reactiveColAvoidance2013,downwash1,downwash3,dcad2019,Engelhardt2016FlatnessbasedCF,navigation2017,reactiveColAvMorgan,reactiveColAvBaca,reactiveColAvMorganJ,speedAdjust2021,gameCollisionAvoidance2020,gameCollisionAvoidance2017,preiss2017,Ferrera2018Decentralized3C,planning2019,preiss2017whitewash,opticalpositioning1,navigate2014} to travel to its coordinate.
With dead reckoning, the wing of the butterfly farthest from (closest to) the origin 
is least (most) accurate because FLSs travel a greater (shorter) distance.  With 5 degrees of error, the butterfly wing farthest from the origin is distorted completely, see Figure~\ref{fig:cmp}.b.
In general, dead reckoning is challenged with realizing symmetrical geometric shapes as simple as a straight line, e.g., the cat's tail which is close to the dispatcher.


The challenge is to design and implement a general purpose framework that localizes FLSs to render any 2D or 3D shape with an arbitrary amount of dead-reckoning error.
This challenge constitutes the focus of this paper.
The ideal framework must be:
\begin{itemize}
\item Fast.
It should localize FLSs as they are deployed.
This results in an incremental rendering where portions of a shape become recognizable as other FLSs are deployed to render the remainder of the shape. 
\item Highly accurate.
A rendering should match the intended point cloud.
With some applications, this may be an overkill because the human eye may not detect one FLS out of its place even with small point clouds consisting of one hundred FLSs, e.g., compare Figure~\ref{fig:cmp}.c with~\ref{fig:cmp}.a.
\item Continuous.
Drones are known to drift~\cite{drift2017,drift2018}.  
FLSs must be able to execute the algorithm continuously to compensate for both their drift and those of their neighbors.

\item Resilient to FLS failures and departures.  FLSs are mechanical devices that may fail~\cite{shahram2021}.  Moreover, their battery has a fixed flight time.
An FLS must be allowed to fly back to a charging station once its battery is depleted per an algorithm such as STAG~\cite{shahram2022}.
The substituting FLS must be able to localize itself.

\item Scalable to render shapes consisting of a large number of FLSs, tens of thousands if not millions.

\item Minimal, requiring as few sensors as possible from the display to enhance its flexibility and reduce its cost.
\end{itemize}

Swarm-Merging, SwarMer, addresses these requirements for 2D and 3D point clouds.  It is designed for an in-door 3D display that lacks line of sight with GPS satellites.  It is a {\em relative} localization system instead of an {\em absolute} one, e.g., the teapot of Figure~\ref{fig:cmp}.c is moved down relative to its ground truth shown in Figure~\ref{fig:cmp}.a.  Given a 2D or 3D point cloud, it approximates the position of the FLSs by forcing the neighboring FLSs to match the relative distance and angle~\cite{deter2022} of their assigned points.  Even though the location of each FLS is not exactly the same as that specified by the point cloud, SwarMer’s relative positioning of FLSs produces shapes true to a point cloud.
In some instances, there is an exact match between the position of FLSs and the coordinates of points in the point cloud.

SwarMer is a decentralized algorithm, requiring each FLS to localize itself relative to another FLS at a time.  Many FLSs may correct their position concurrently and independently.  Hence, SwarMer may render a portion of a shape while FLSs are being deployed to complete the rest of the shape.
This makes SwarMer fast.

SwarMer constructs swarms~\cite{reynolds87} of FLSs.
Movement of one FLS causes all FLSs in its swarm to move.
SwarMer includes provisions that prevent 
isolated FLSs or fragmented swarms when there should be one swarm.
These provisions make SwarMer resilient to FLSs failing and departing to charge their battery. 


SwarMer requires FLS to use dead reckoning to position themselves relative to one another repeatedly.  
Each time, depending on the relative localization technique of the FLSs, the distance traveled by an FLS decreases. 
Hence, error with inter-FLS dead-reckoning decreases, enhancing SwarMer's accuracy.
See Figure~\ref{fig:cmp}.c.


Figures~\ref{fig:cmp}.d and~\ref{fig:cmp}.e show the use of triangulation and trilateration to localize FLSs.
Both are designed for use with anchors that have precise positions~\cite{uwb2015,zebra,chorus2019,uwb2022,uwbfusion2022}, e.g., GPS satellites~\cite{gps2008} used by GPS enabled drones for outdoor light shows.
In Figures~\ref{fig:cmp}.d and~\ref{fig:cmp}.e, an FLS lacks line of sight with a GPS satellite.
It uses those FLSs with a high confidence in their location as its anchor with both triangulation and trilateration.
The position of these anchors is not 100\% precise.
This causes triangulation and trilateration to produce distorted shapes. 



\noindent {\bf Assumptions} of the SwarMer framework about its FLSs are as follows.
First, FLSs are cooperative.
Second, an FLS may adjust its radio range by manipulating its transmission power to communicate with FLSs that are farther away~\cite{sspower}.
Third, each FLS has a unique id, FID.
FID may be assigned by the FLS hardware, e.g., MAC address of its networking card.
Alternatively, a dispatcher~\cite{shahram2022} may assign an FID to an FLS as it deploys FLSs.
Fourth, an FLS knows its assigned coordinate (a point in a point cloud) and the coordinates of its neighbors.
The exact number of neighbors known by an FLS is a tradeoff between the amount of storage required from each FLS and the required network bandwidth for the FLSs to communicate and obtain this information from one another.

\noindent {\bf Contributions} of this paper are:
\begin{itemize}
    \item SwarMer as a relative localization framework to render 2D and 3D shapes.  This decentralized framework uses nature inspired swarm concepts.  
    FLSs may execute SwarMer continuously.
    To the best of our knowledge SwarMer is novel and has not been published elsewhere.  (See Section~\ref{sec:sm}.)
    \item A localization technique is a plugin for the SwarMer framework.  We describe two alternative localization techniques using signal strength and physical movement of FLSs relative to one another.  In addition, we quantify the tradeoffs associated with these techniques.  (See Section~\ref{sec:localize} and \url{https://youtu.be/Lh11tWWOP5Y}.)

\item Triangulation and trilateration\footnote{Both triangulation and trilateration may serve as a localization plugin of the SwarMer framework.  We do not pursue them in this role because the signal strength technique of Section~\ref{sec:localize} is simpler and highly accurate.} as comparison yardstick to evaluate SwarMer.
Obtained results demonstrate the superiority of the SwarMer framework, compare Figure~\ref{fig:cmp}.c with~\ref{fig:cmp}.d and e.
(See Section~\ref{sec:cmp}.)

    \item We present a simulation and an implementation of SwarMer.  The MATLAB simulator is used to study and obtain insights into the alternative design decisions.  The most promising designs with adjustable configuration parameters are implemented using the Python 3.9 programming language with UDP sockets.  We are actively exploring their deployment in a swarm of drones.
    (See Section~\ref{sec:impl}.)
    \item We open source our software implementations, and the 2D and 3D data used in this study for further use and investigation by the scientific community.  See \url{https://github.com/flyinglightspeck/SwarMer} for details. 

    \item A limitation of SwarMer is that it may shift a shape along different dimensions.
    It requires an accurate reference point or an {\em oracle} FLS to address this limitation.
    (See Section~\ref{sec:random}.)
 
\end{itemize}

The rest of this paper is organized as follows.
Section~\ref{sec:terminology} provides the terminology used by this paper and a formal statement of the localization problem.
Section~\ref{sec:sm} presents SwarMer.
We evaluate SwarMer using 2D and 3D point clouds in Section~\ref{sec:eval}.
Section~\ref{sec:cmp} provides our Matlab implementation of triangulation and trilateration, comparing them with SwarMer.
Section~\ref{sec:impl} presents an emulation of SwarMer.
Related work are described in Section~\ref{sec:related}.
Section~\ref{sec:future} presents our future research directions.

\section{Terminology, Problem Statement}\label{sec:terminology}
Given the novelty of FLSs and a 3D FLS display, this section introduces terminology for concepts in order to state the problem solved by SwarMer. 
We use this terminology throughout the paper.  
The SwarMer framework is independent of the assumed terminology and may use alternative ontologies that match the concepts per definitions provided in this section.

To render an illumination, an FLS display constructs a 3D mesh on the display volume. 
A cell of this mesh, a display cell~\cite{shahram2022}, is dictated by the downwash of an FLS~\cite{dcad2019,preiss2017,downwash3,Ferrera2018Decentralized3C,planning2019}.  
Assuming an FLS is a quadrotar, a cell may be an 
ellipsoid~\cite{dcad2019,preiss2017,downwash3} or a
cylinders~\cite{Ferrera2018Decentralized3C,planning2019} that 
results in a larger separation along the height dimension.
Each display cell has a unique Length, Height, and Depth (L,H,D) coordinate~\cite{shahram2022}.
We use the L, H, D coordinate system instead of X,
Y, Z to identify a cell because there is no consensus on one definition of the Y and
Z axes. While the picture industry uses the Z axis as the depth,
mathematicians use the Y axis for the depth. 
It is trivial to map our
L, H, D coordinate system to either definition without ambiguity.



We assume a fixed coordinate system for the FLS display with (0, 0, 0) as its bottom left corner.
Moreover, we assume the geometry of a 3D point cloud identifies the location of a display cell in an illumination cell.
This is the
(L,H,D) coordinate of a point assigned to an FLS to illuminate.

\begin{definition}
{\em Ground truth} is the ideal location and orientation of an FLS in an FLS display.  It is dictated by the geometry of a point cloud, the (L,H,D) coordinates of a point.  An FLS is assigned the coordinates of a point in the ground truth to illuminate.
\end{definition}

\begin{definition}
{\em Estimated truth} is the actual location and orientation of an FLS in an FLS display.
It is an FLS's estimate of the ground truth and may match the ground truth with some degree of accuracy including perfect (100\%) accuracy.
\end{definition}

\begin{definition}
A {\em vector} $\vec{i,j}$ exists between any two points $i$ and $j$ in a point cloud.
It has a fixed length $|\vec{i,j}|$.
With 2D point clouds, it has an angle $\alpha$ relative to the horizontal L-axis.
With 3D point clouds, in addition to $\alpha$, it has an angle $\theta$ relative to the height, H-axis, of a display.
See Figure~\ref{fig:locss3D}.
\end{definition}


\begin{problem}\label{prob:hausdorff}
Given a 2D (3D) point cloud with each point assigned to a unique FLS, localize FLSs to minimize 
the Hausdorff distance~\cite{Huttenlocher93} between the estimated truth and the ground truth.
\end{problem}

Section~\ref{sec:hd} provides a formal definition of how we compute the Hausdorff distance~\cite{Huttenlocher93}.
Hausdorff distance is used to quantify SwarMer's accuracy and SwarMer is independent of it.

\subsection{Hausdorff Distance, HD}\label{sec:hd}
We use the Hausdorff distance~\cite{Huttenlocher93} to compare the FLS rendering E, i.e., the estimated truth, with the ground truth G, Hausdorff(E,G) or HD for short.
This metric reports the maximum error in distance between E and G after applying a translation.
A lower value is better with zero reflecting a perfect match between E and G.
Below we describe how we compute the maximum error and our translation process in turn.

To compute the maximum error between E and G, for each coordinate $e_i$ in E, we find the minimum distance between $e_i$ and the coordinate $g_j$ of each point $j$ in G. 
Let \{EG\} denote this set of values.
Next, for the coordinate of each point $g_i$ in G, we compute the minimum distance between $g_i$ and the coordinate of each FLS $e_j$ in E.
Let \{GE\} denote this set of values.
The Hausdorff distance is the maximum of the values in these two sets, HD = max(max(\{EG\}), max(\{GE\})).

The translation of E to G with the MATLAB simulator of Section~\ref{sec:eval} is different than with the implementation of Section~\ref{sec:impl}.  
With the simulator, we translate E to G using the following stochastic technique.
We select $\{R\}$ random FLSs from E; $|R|$=90 FLSs in all our experiments.
For each FLS $i$ in $\{R\}$, we identify its assigned point in G.
We compute the vector $\vec{D_i}$ that reduces the distance between the pair to zero. 
We apply $\vec{D_i}$ to the $\{R\}$ FLSs and compute the sum of errors between the $\{R\}$ FLSs and their assigned points in G, $E_i=\sum_{j=1}^{|R|} e_j-g_j$.
We identify the FLS $k$ with minimum error\footnote{In case of ties we pick one randomly.} $E_k$ ($1 \leq k \leq |R|$).
Our translation process applies
vector $\vec{D_k}$ to all FLSs in E.
Subsequently, we compute HD.
The complexity of this stochastic technique is $90 \times (90 + 3 \times 90) + F$ since it applies each translation to 90 points, computes the distance between each pair of 90 corresponding points, and sums them. Then applies the one with the minimum sum of distances to the $F$ FLSs.
More generally, the complexity is O($c \times |R|^2 + F$) where c is a constant.

With the implementation, we compute the center\footnote{The center of a point cloud is the average of the coordinates of its points.} of each point cloud $E$ and $G$, $C_E$ and $C_G$, respectively.
We compute a vector $\vec{D}=C_G-C_E$
We add all $\vec{D}$ to each point in the point cloud E.
Finally, we compute HD.
The complexity of computing the center of each point cloud is O($F$) where F is number of FLSs. 
With the additional scan of E to apply $\vec{D}$, the overall complexity is O($3F$).

The translation process is required because SwarMer is a relative localization technique.

\section{Swarm-Merging (SwarMer)}\label{sec:sm}

SwarMer is a decentralized algorithm that assumes the error in dead reckoning decreases when FLSs travel shorter distances.
A dispatched FLS travels to its assigned destination using dead reckoning.
When it arrives at its destination, 
SwarMer requires the FLS to localize relative to its neighbors using dead reckoning repeatedly.
This reduces the travelled distance to enhance the accuracy of dead reckoning, enabling the FLS to approximate the ground truth more accurately.

As soon as an FLS $F_i$ arrives at its destination, it constructs a swarm consisting of itself with its Swarm-ID set to its FID, $\xi_i$=$i$.
Next, $F_i$ expands its radio range to identify an FLS $F_j$ with a different swarm-id, $\xi_i \neq \xi_j$.
$F_i$ challenges $F_j$ to merge into one swarm. 
Assuming $F_j$ accepts the challenge,
the one with a lower Swarm-ID becomes the {\em anchor} FLS and the other becomes the {\em localizing} FLS.
Assuming $\xi_i < \xi_j$, 
the localizing FLS, $F_j$, computes a vector $\vec{V}$ for its movement relative to its anchor $F_i$ to better approximate the {\em ground truth}. Section~\ref{sec:localize} describes two techniques to compute $\vec{V}$.

An FLS maintains a binary status property initialized to \texttt{available}.  Its other possible value is \texttt{busy}.
When $F_j$ is challenged and its status is \texttt{available} then it changes its status to \texttt{busy} and notifies the FLSs that constitute its swarm to change their status to \texttt{busy}.
$F_i$ also changes its status to \texttt{busy} and informs those FLSs in its swarm to change their status to \texttt{busy}.
Each message includes whether the swarm is localizing or anchor.
$F_i$ and $F_j$ determine this independently, by comparing their Swarm-ID, $\xi_i$ and $\xi_j$.
With $\xi_i<\xi_j$, in addition to the \texttt{busy} status change, $F_i$'s ($F_j$'s) message includes the anchor (localizing) flag.

An \texttt{available} FLS accepts a challenge always.  A \texttt{busy} FLS accepts a challenge only if it is an anchor and its Swarm-ID is lower than that of a challenger.  This FLS will serve as the anchor for the challenger.
An FLS may serve as the anchor for multiple localizing FLSs beloning to different swarms.
Moreover, different FLSs belonging to the same swarm may serve as anchors for multiple localizing FLSs belonging to different swarms.
Hence, SwarMer may merge two or more swarms ($M \geq 2$) into one. 

Once $F_j$ computes a vector $\vec{V}$ for its movement relative to $F_i$, $F_j$ informs members of its swarm to 
change their Swarm-ID to $F_i$'s Swarm-ID ($\xi_i$) and move along $\vec{V}$.
$F_j$ changes its Swarm-ID to $F_i$'s Swarm-ID ($\xi_i$) and moves along $\vec{V}$ to localize itself relative to $F_i$.
Next, it changes its status from \texttt{busy} to \texttt{available}. 
$F_j$'s local movement and status change may occur concurrently with its message transmission to the members of its swarm.  
Members of $F_j$'s swarm change their Swarm-ID to $\xi_i$, move along $\vec{V}$, and change their status to \texttt{available} independently.
Finally, $F_j$ notifies $F_i$ to unanchor itself, enabling it to localize relative to other FLSs.
This completes merging of the $F_i$ and $F_j$ swarms into one with the participating FLS's Swarm-ID set to $\xi_i$.



SwarMer localizes FLSs using their nearest neighbor by requiring an FLS to expand its radio range.
FLSs must measure distance and angle relative to one another to localize. 


\begin{definition}\label{def:swarm}
A {\em swarm} consists of a collection of FLSs with the same Swarm-ID $\xi$.
$\xi$ is set to the minimum FID of the FLSs in the swarm.
FLSs that constitute a swarm move together~\cite{boids87,bswarm2015}.
\end{definition}


\begin{definition}
A {\em localizing} FLS has a Swarm-ID greater than its anchor FLS.  It informs all members of its swarm to change their status to \texttt{busy}, preventing them from localizing relative to a different swarm.
It computes a new coordinate and orientation $\phi$ relative to its anchor FLS.
Both are in the FLS's estimated truth.
Using its current and new coordinates, it computes a vector $\vec{V}$ to travel to its new coordinate.
The FLS informs all members of its swarm to change their Swarm-ID to that of its anchor FLS, move along $\vec{V}$ using dead reckoning, and orient themselves using $\phi$.
The FLS changes its Swarm-ID to that of the localizing FLS, uses dead reckoning to travel along vector $\vec{V}$ to its new coordinate, and orients itself using angle $\phi$.
The localizing FLS and members of its swarm move
along $\vec{V}$ as long as $\vec{V}$'s magnitude exceeds a threshold. 
The localizing FLS changes its status to \texttt{available} and notifies the anchor FLSs to change its status to \texttt{available}.
\end{definition}

\begin{definition}
An {\em anchor} FLS remains stationary and changes its status to \texttt{busy}, enabling
one or more localizing FLSs belonging to different swarms to compute a vector $\vec{V}$ and orientation $\phi$ relative to it.
It has a Swarm-ID lower than its localizing FLSs.
It notifies members of its swarm to change their status to \texttt{busy}, preventing them from localizing.

\end{definition}

\begin{figure}
\centering
\includegraphics[width=3.5in]{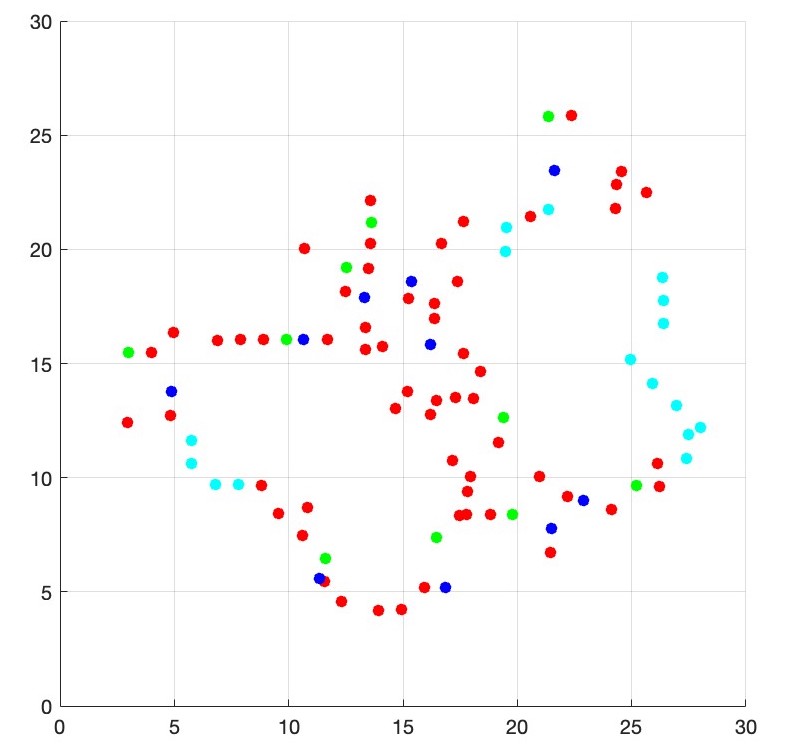}\hfill
\caption{
Greens and dark blues are localizing and anchored FLSs, respectively.
Members of their swarms are in red.  Available FLSs are in light blue.
}
\label{fig:concurrent}
\end{figure}


FLSs belonging to different swarms may localize concurrently.
One or more FLSs of a swarm may serve as the anchor for different localizing FLSs.
Figure~\ref{fig:concurrent} shows 10 localizing FLSs (green), their anchored FLSs (dark blue), members of their swarms (red), and available FLSs (light blue).


When an FLS expands its radio range, it may encounter $M-1$ FLSs with different Swarm-IDs.
The FLS with the lowest Swarm-ID becomes the anchor while $M-1$ FLSs localize relative to it.  
This merges $M$ swarms into one.
SwarMer may limit the value of M to 2 or higher ($\infty$), e.g., 2 for binary merging of swarms\footnote{Section~\ref{sec:numswarms} shows $M$=$\infty$ renders a shape accurately faster when compared with $M$=2.}.
The value $\infty$ merges as many candidates as possible.

SwarMer may use different techniques to select the identity of the anchor swarm.
Thus far, we have described the {\em Lowest Swarm-ID}, a simple technique that uses the FLS with the lowest Swarm-ID as the anchor.  
A more complex technique is to use the FLS belonging to the swarm with the most number of FLSs as the anchor FLS.
It may require an FLS to use 
a decentralized peer-to-peer technique to estimate the size of its swarm~\cite{p2psize2003}.
In case of ties with multiple swarms having the same size, the one with the lowest Swarm-ID may serve as the anchor.
Section~\ref{sec:anchortech} evaluates four techniques.
It shows while one technique may be superior for a specific point cloud, no technique is superior across all point clouds.
Hence, we use the lowest Swarm-ID for the rest of this section.

With $F$ FLSs, the number of rounds performed by SwarMer to construct one swarm is $log~F$.
With $M$=2, the number of rounds is deterministic.
With $M$=$\infty$, the base of the log is a function of the average number of swarms merged in a round.
It depends on the density of the point cloud and how many swarms are merged at a time.
See discussions of Figure~\ref{fig:sharedanchors}.

Multiple FLSs, say $\mu$, of a swarm may race with one another to localize relative to different FLSs belonging to different anchor swarms concurrently.
This results in a race condition between $\mu$ localizing FLSs generating 
different vectors for the FLSs of the localizing swarm.
SwarMer ensures FLSs of the localizing swarm applies at most one vector, discarding $\mu-1$ vectors.
This partitions the localizing swarm into $\mu$ swarms.
A localizing FLS that uses anchor FLS $F_i$ generates a message containing its Old Swarm-ID ($\xi_{O}$), New Swarm-ID ($\xi_{Ni}$), and its computed vector.
There are $\mu-1$ messages with different $\xi_{Ni} < \xi_{O}$ racing with one another.
A receiving FLS compares its Swarm-ID with $\xi_{O}$ contained in the message.
If they are equal then it sets its Swarm-ID to $\xi_{Ni}$ and travels along the provided vector in the message.
This FLS will discard the $\mu-1$ messages since its Swarm-ID $\xi_{Ni}$ does not equal to the old Swarm-ID of the messages.
The consequence is that the localizing swarm is fragmented into $\mu$ swarms.
These swarms will subsequently localize relative to one another and merge into one.

To illustrate, assume FLSs A and B of Swarm 3 localize relative to an FLS of two different swarms with Swarm-IDs 1 and 2.  FLSs A and B issue a Busy message to all FLSs that constitute Swarm 3.  After localizing, FLSs A and B generate a message with their independently computed vector, $\xi_{O}$=3, and the new Swarm-ID $\xi_{N}$=\{1 or 2\}.  These messages race with one another.  A receiving busy FLS $F_k$ with Swarm-ID $\xi_{k}$=3 changes its Swarm-ID to the Swarm Id of the winner message (say 1, $\xi_{k}$=1).
$F_k$'s swarm-ID $\xi_{k}$=1 does not match the loser’s message with $\xi_{O}$=3, causing $F_k$ to discard this message.
Thus, one vector is applied by a unique and distinct subset of FLSs that constitute Swarm 3, fragmenting Swarm 3 into two swarms with two different Swarm IDs, 1 and 2.  

As swarms merge and form larger swarms, SwarMer saves battery power of FLSs by requiring only those FLSs with a missing neighbor to expand their radio range.
This saves battery power of many FLSs by preventing them from expanding their radio range only to discover FLSs that decline their challenge to merge.
FLSs with no missing neighbors are termed Relatively-Complete, R-Complete.  
They wait to be challenged.

\begin{definition}
An FLS is {\em Relatively Complete, R-Complete}, once its expands its radio range and discovers $\eta$ FLSs in its estimated truth that match its ground truth and have its Swarm-ID, i.e., they are a part of the same swarm.
An R-Complete FLS does not expand its radio range and does not challenge other FLSs to join its swarm.
The value of $\eta$ is adjustable and maybe customized for efficient rendering of a point cloud.
\end{definition}

SwarMer imposes the following requirement on all FLSs.
A swarm consisting of an FLS
with busy neighbors belonging to a different swarm is required to localize itself relative to one of the neighbors and join its swarm.
This requirement makes SwarMer resilient to FLS failures and FLSs leaving to charge their battery.
In both cases, a new FLS arrives to substitute for the departing FLS.
The requirement causes this new FLS to localize and join the existing swarm immediately.
In addition, this requirement prevents formation of multiple disjoint swarms that should be one with small values of $\eta$.

One may configure SwarMer to localize FLSs continuously.
Once all FLSs constitute one swarm, they may thaw their swarm membership.
Each FLS resets its Swarm-ID to its FID, Swarm-ID=FID.
This forms swarms consisting of one FLS without impacting the position of each FLS.
Next, the FLSs repeat the localization process of SwarMer.  
Section~\ref{sec:evalres} shows repeated application of SwarMer enhances the quality of a rendering, i.e., lowers the Hausdorff distance.
This repetition is fast.
Section~\ref{sec:impl} shows SwarMer drops the Hausdorff distance to a value below 0.01 after a few repetitions in tens of seconds.

There are multiple ways of thawing the final swarm in a decentralized manner.
One approach may require an FLSs to periodically estimate the size of its swarm using a decentralized counting technique such as~\cite{p2psize2003}.
Each FLS may be provided with the size of the point cloud when it is deployed.
Once the two numbers are approximately the same then the FLS broadcasts a thaw message.
All FLSs that receive the thaw message reset their SwarmID to their respective FID and repeat the thaw message with some probability.

An alternative approach may require those FLSs at the farthest corner of an illumination to set a timer once they are R-Complete.
When this timer expires, an FLS expands its radio range to the maximum possible and generates a thaw message, announcing its Swarm-ID.
If a receiving FLS has a different Swarm-ID then it broadcasts a message to abort swarm thawing, causing the transmitter to sleep and try again.
Otherwise, it repeats the thaw message using its Swarm-ID with some probability. 
Once an FLS receives a fixed number of thaw messages, it sets its Swarm-ID to its FID and repeats the localization process using SwarMer.
An evaluation of these alternatives is a future research direction.

\subsection{Relative Localization}\label{sec:localize}
This section describes two techniques for a localizing FLS to position itself relative to an anchor FLS.
They are plugins for the SwarMer framework and may be implemented using a variety of techniques described in Section~\ref{sec:implSS}.

\begin{figure}
\centering
\includegraphics[width=3.5in]{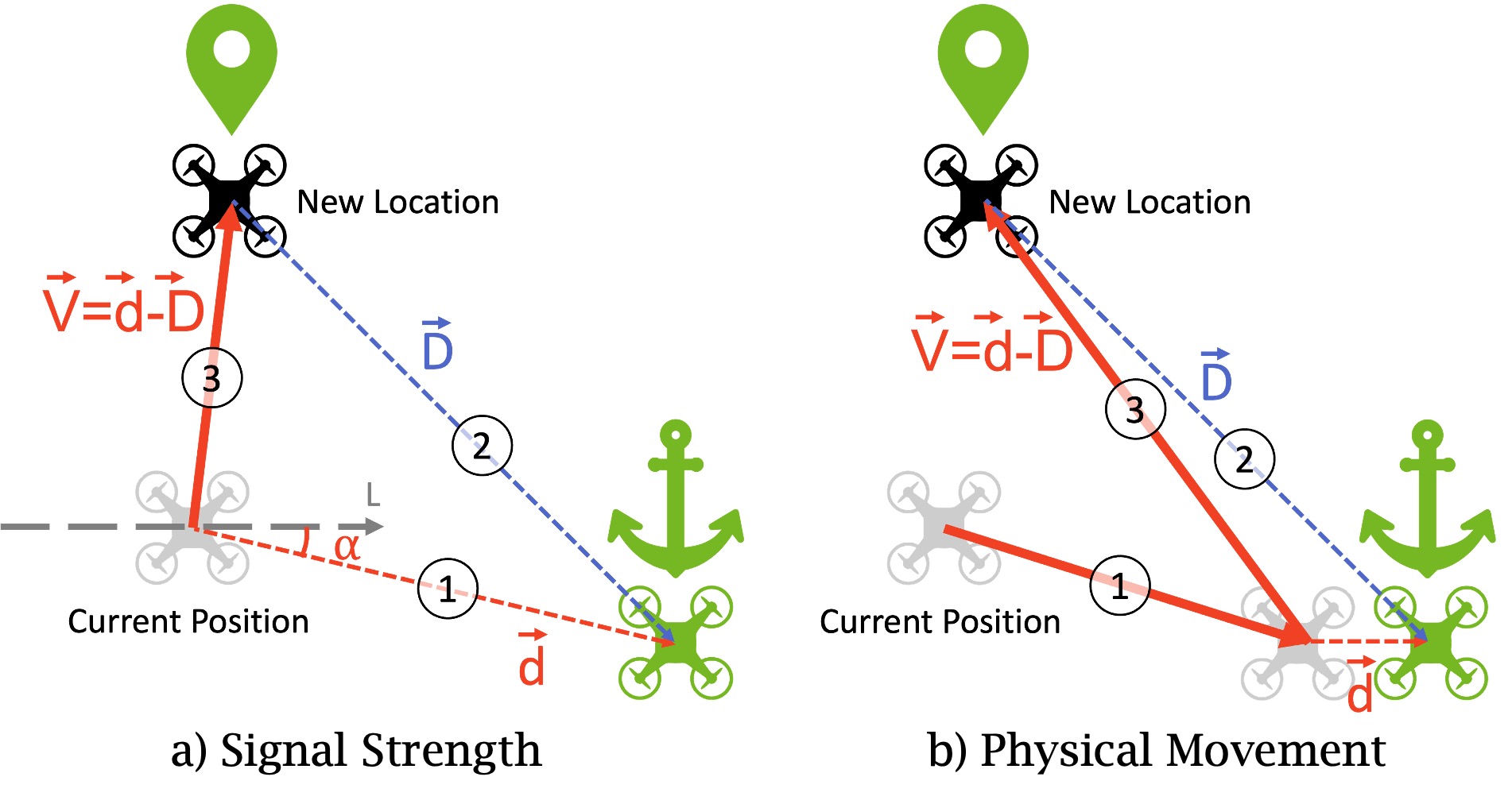}\hfill
\caption{SwarMer FLS Localization in 2D.}
\label{fig:locss}
\end{figure}

\begin{figure}
\centering
\includegraphics[width=3.5in]{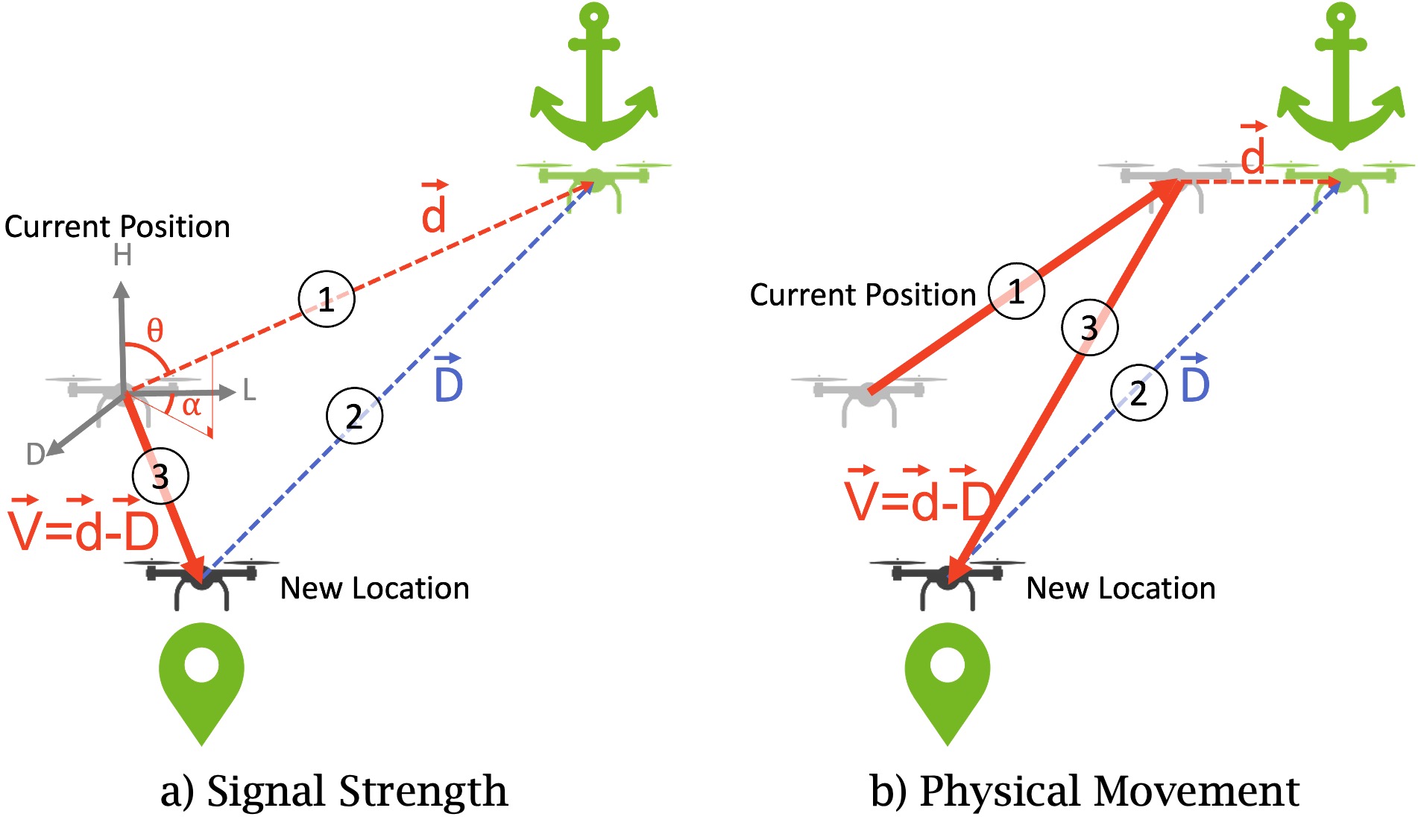}\hfill
\caption{SwarMer FLS Localization in 3D.}
\label{fig:locss3D}
\end{figure}

The first technique, Signal Strength (SS), assumes an FLS (a) has sensors to detect both the angle and the strength of the signal received from another FLS,
and (b) may convert these sensor readings to distance from the FLS generating the signal.
One may consider and evaluate alternative sensors when designing and implementing SS with FLSs.  
Bluetooth~\cite{bluetooth1,bluetooth2}, Wi-Fi~\cite{wifi1,wifi2}, RFID~\cite{rfid1,rfid2,rfid3,rfid4}, UWB~\cite{uwb1,uwb2,uwb3}, Lidar~\cite{lidar1,rgblidar}, RGB cameras~\cite{rgb1,rgb2,rgb3,rgblidar} and infrared~\cite{opticalpositioning1,opticalpositioning2,preiss2017} may be used to obtain distance information.
The angle information may be obtained by mounting an antenna array on an FLS, e.g., 
Bluetooth~\cite{ble21}, Wi-Fi~\cite{caliberate21}.
Angle of Arrival, AOA~\cite{aoa2013}, obtains the angles at which the signal sent by a localizing FLS reaches the antennas mounted on an anchor FLS.
Several surveys document the alternative sensor technologies, their strengths and weaknesses~\cite{chenSurvey2022,gusurvey2009,luca2014,zafari2019}.

In 2D, SS requires the localizing FLS to request the anchor FLS to generate a signal.
It uses its sensors to estimate vector $\vec{d}$.  This vector is defined using angle $\alpha$ relative to the L-axis with distance |d| from the anchor FLS, see Step 1 in Figure~\ref{fig:locss}.a.
Next, the localizing FLS uses the ground truth to calculate vector $\vec{D}$.
The tail (head) of $\vec{D}$ is the position of the localizing (anchor) FLS in the ground truth, see Step 2 in Figure~\ref{fig:locss}.a.
We subtract the two vectors to obtain $\vec{V}$, $\vec{V}=\vec{d}-\vec{D}$.
The localizing FLS moves along vector $\vec{V}$ to its new position, see Step 3 in Figure~\ref{fig:locss}.a.

The second technique, named Physical Movement (PM), requires the localizing FLS to position itself next to the anchor FLS using its Inertial-Measurement-Unit, IMU.
Step 1 of Figure~\ref{fig:locss}.b shows this step in 2D.
Next, the localizing FLS computes its angle and distance from the anchor FLS in the ground truth, vector $\vec{D}$ of Step 2 in Figure~\ref{fig:locss}.b.
It also computes its displacement vector relative to the anchor, Vector $\vec{d}$.
We subtract the two vectors to obtain $\vec{V}$, $\vec{V}=\vec{d}-\vec{D}$.
The localizing FLS moves along vector $\vec{V}$ to its new position, see Step 3 in Figure~\ref{fig:locss}.b.

In 3D, SS must also compute the angle $\theta$ relative to the height H, see Figure~\ref{fig:locss3D}.a.  PM uses a similar angle implicitly (not shown in Figure~\ref{fig:locss3D}.b) as a part of its Step 1.

\begin{figure}
\centering
\includegraphics[width=3.4in]{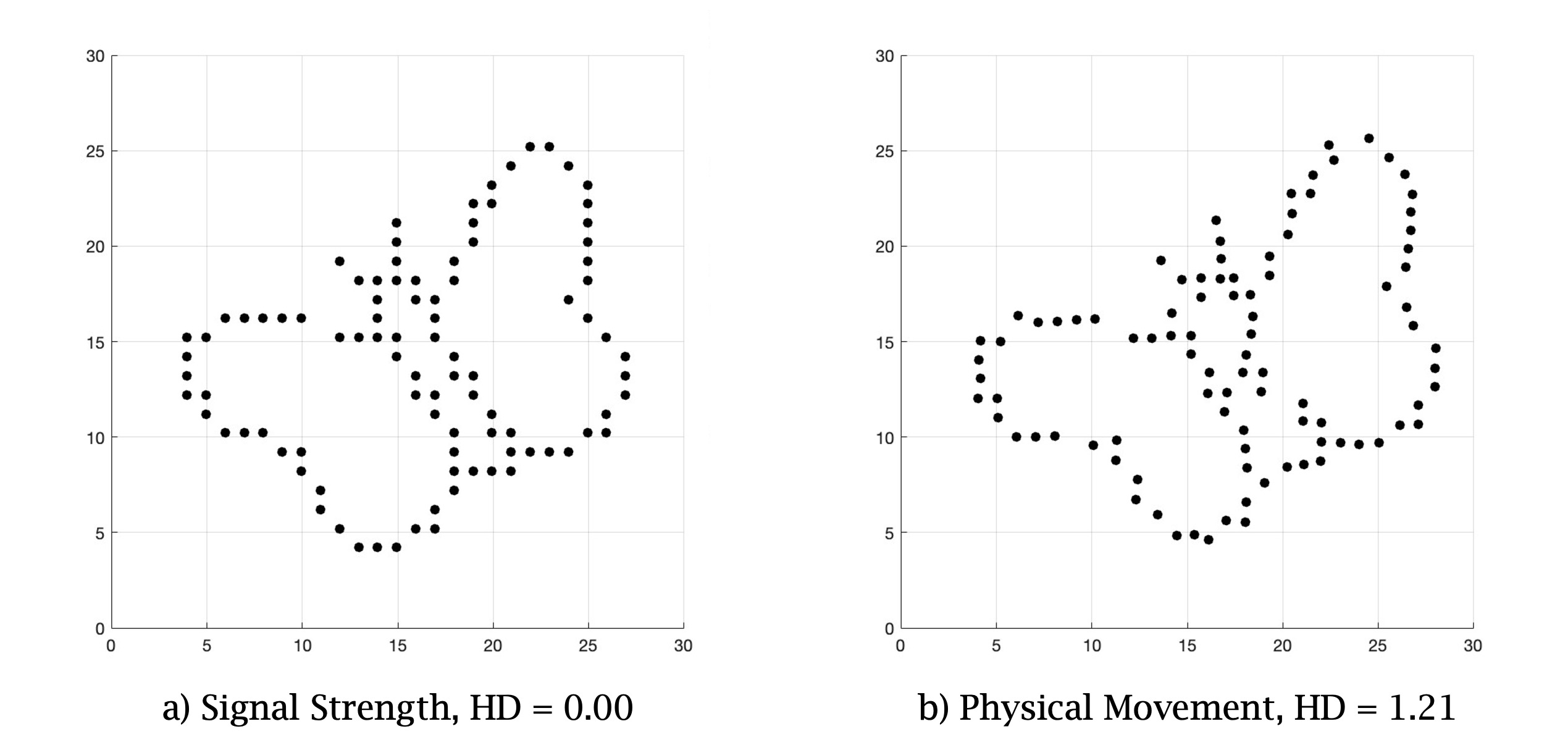}\hfill
\caption{A comparison of SS with PM, $\epsilon$=5$\degree$.}
\label{fig:ssVSpmI}
\end{figure}

\begin{figure}
\centering
\includegraphics[width=3.65in]{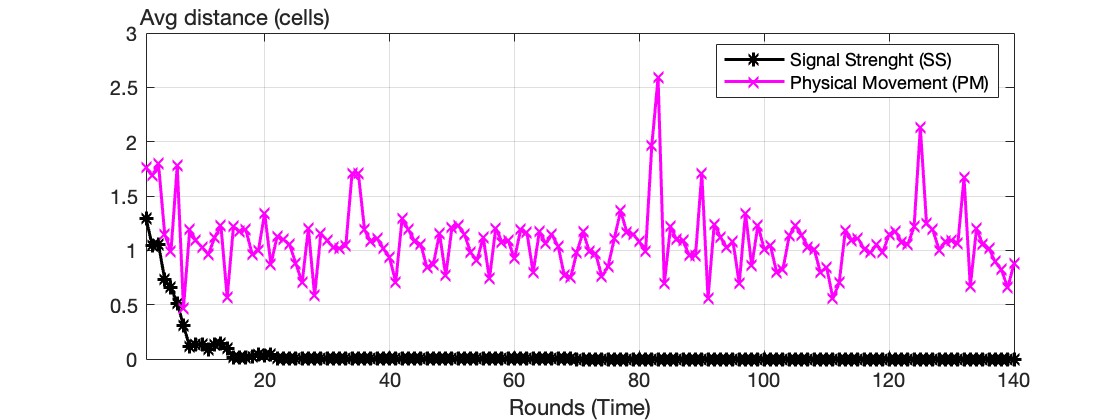}\hfill
\caption{Dead reckoning distance for the 2D Butterfly of Figure~\ref{fig:ssVSpmI}.}
\label{fig:ssVSpmDead}
\end{figure}

\begin{figure}
\centering
\includegraphics[width=3.65in]{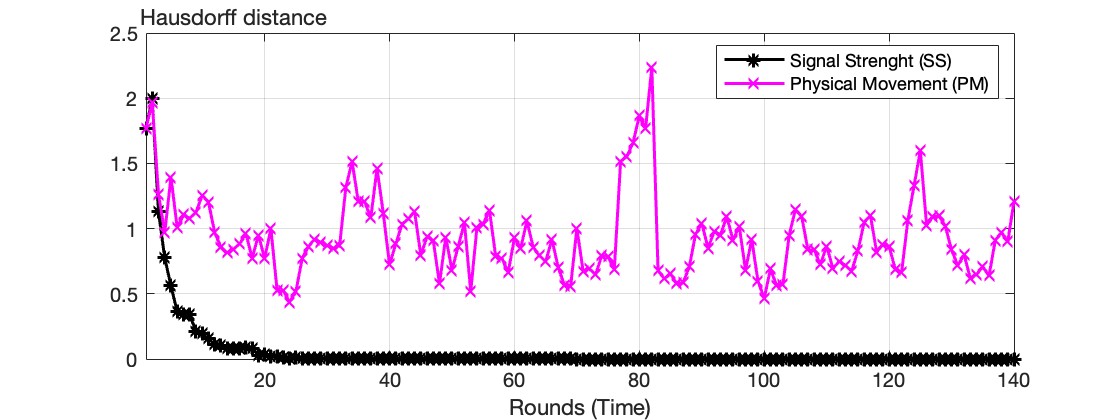}\hfill
\caption{Hausdorff distance for the 2D Butterfly of Figure~\ref{fig:ssVSpmI}.}
\label{fig:ssVSpmH}
\end{figure}

Generally, the dead reckoning distance travelled by a localizing FLS is greater with PM when compared with SS.
This results in a higher error with PM, enabling SS to provide more accurate renderings.
To illustrate, Figure~\ref{fig:ssVSpmDead} shows the dead reckoning distance with the 2D butterfly.
Generally, SS reduces this distance somewhere between a factor of 5x to 10x with 2D shapes.
This increases the accuracy of the rendered shapes, see Figure~\ref{fig:ssVSpmI}.a. 
Figure~\ref{fig:ssVSpmH} shows the Hausdorff distance for the two alternatives with the 2D butterfly.
The Hausdorff distance decreases close to zero with SS because the dead reckoning distance is decreasing as a function of the number of iterations, see Figure~\ref{fig:ssVSpmDead}. 
PM does not realize the same.

The x-axis of Figures~\ref{fig:ssVSpmDead} and~\ref{fig:ssVSpmH} is an estimate of time.
SS approximates the ground truth faster than PM.

\begin{figure}
\centering
\includegraphics[width=3.4in]{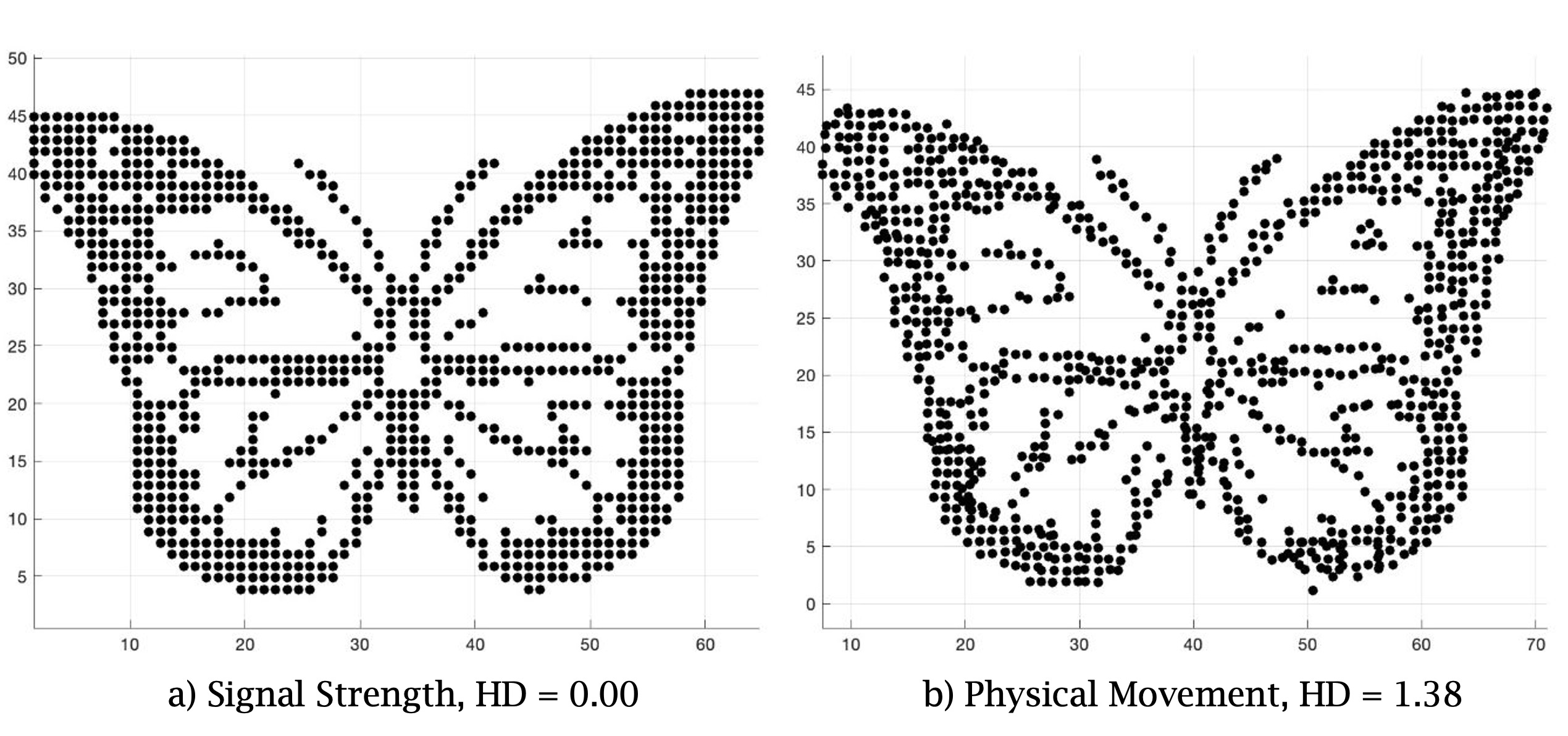}\hfill
\caption{A comparison of SS with PM, $\epsilon$=5$\degree$.}
\label{fig:ssVSpmBig}
\end{figure}
\begin{figure}
\centering
\includegraphics[width=3.65in]{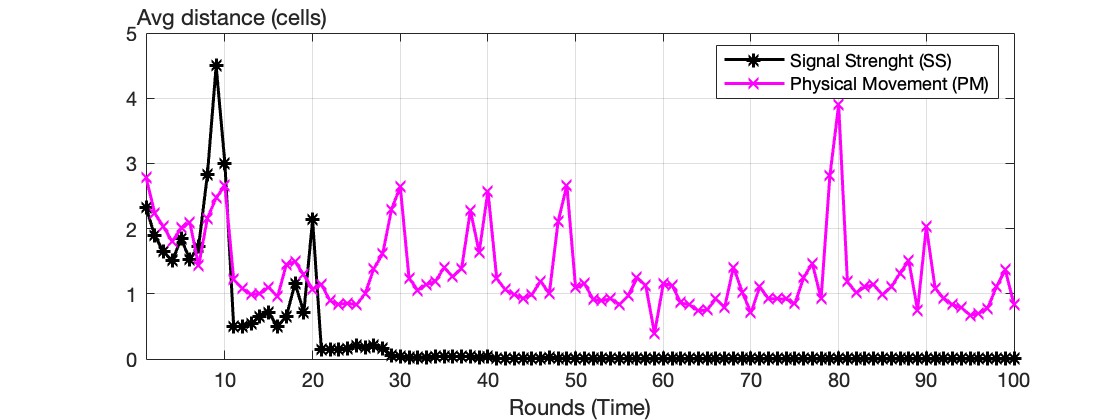}\hfill
\caption{Dead reckoning distance for the 2D Butterfly of Figure~\ref{fig:ssVSpmBig}.}
\label{fig:ssVSpmDeadBig}
\end{figure}
\begin{figure}
\centering
\includegraphics[width=3.65in]{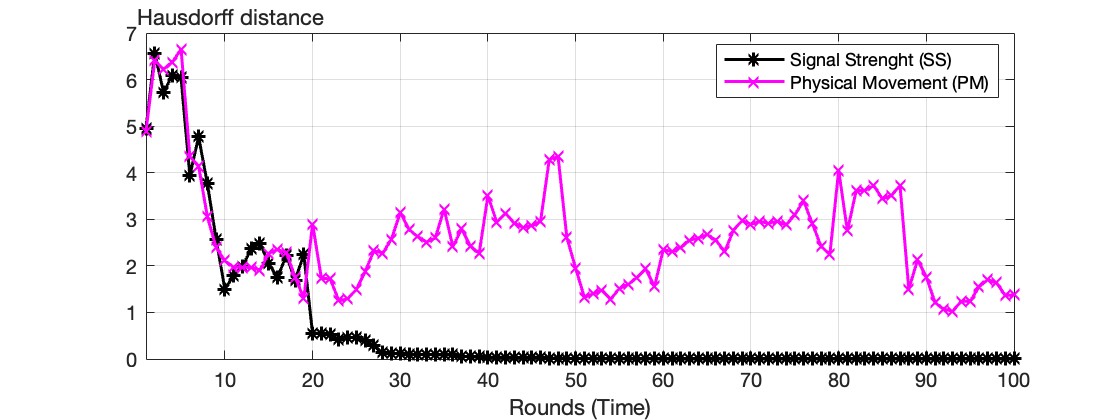}\hfill
\caption{Hausdorff distance for the 2D Butterfly of Figure~\ref{fig:ssVSpmBig}.}
\label{fig:ssVSpmHBig}
\end{figure}

With SS, the average dead reckoning distance and Hausdorff distance do not decrease monotonically as a function of the rounds.
Figures~\ref{fig:ssVSpmDeadBig} and~\ref{fig:ssVSpmHBig} show this for the 2D butterfly of Figure~\ref{fig:ssVSpmBig} consisting of 1008 FLSs.
This shape requires FLSs to be aligned in straight lines.
SS is more effective than PM in realizing these lines, compare Figure~\ref{fig:ssVSpmBig}.a with~\ref{fig:ssVSpmBig}.b.

In summary, SS is both faster and more accurate than PM.  It is also more energy efficient by minimizing the distance traveled by FLSs.  This maximizes the flight time of an FLS on a fully charged battery.  These observations hold true with both 2D and 3D shapes.

\subsubsection{Possible Implementations of SS}\label{sec:implSS}
Relative state estimation techniques have been an active area of research and they may be the basis of an SS implementation.
Some studies use radio technology such as Ultra-WideBand (UWB) to measure distance to estimate the relative 3-DOF location between UAVs~\cite{uwb2015,uwb2016,snaploc2019}.
They report measuring distances exceeding 5-7 cm accurately~\cite{snaploc2019,uwb2015}.  
Some techniques use the odometry information to refine UWB measurements~\cite{uwbcao2021}.

Visual odometry techniques~\cite{vision2007,directodometry2018,markerless2010,BABINEC20141} will equip each FLS with an onboard camera to observe other FLSs in its field of view (FoV).
They require a texture rich and a well-lit environment. 
Their accuracy is dictated by the FoV of the on-board camera~\cite{xu2021,xu2022}.

Several studies synergize the strength of different sensors by fusing their measurements to enhance accuracy.
There are techniques to integrate WiFi access points\footnote{WiSion reports position error of 35.25 cm, attitude error of 2.6$\degree$ with a maximum linear velocity of 1.7 m/s~\cite{wision2023}.} 
with IMU~\cite{md-track2018,wision2023}, IMU with visual features from a camera~\cite{orb2017,vins2018,openvins2020}, LiDAR with odometry~\cite{lidar2017,loam2014,floam2021,loamlivox2020}, UWB with IMU and images from a camera~\cite{xu2021,xu2022,viunet2023,uwbIMUvisual2021}, UWB with IMU and LiDAR\footnote{LIRO uses a different number of UWB anchors.  With 3 anchors, it measures distances with 15 to 21 cm accuracy.  It reports angle of rotation with 0.6 to 1.6 degrees of accuracy.}~\cite{liro2021}.
Laser-based approaches such as LiDaR are highly accurate.  However, they may not be suitable for use with FLSs due to their required power, weight, and cost~\cite{wision2023}.

To evaluate a candidate technology, one must evaluate their scalability to thousands and millions of FLSs to illuminate 2D and 3D shapes.

\subsection{FLS Failure and Battery Charging}\label{sec:leases}
An FLS is a mechanical device that may fail.
One may use the concept of leases~\cite{gray89,iq14} to prevent 
failure of a localizing FLS from causing another FLS to remain anchored indefinitely.
When a localizing FLS requests another FLS to serve as its anchor, the anchored FLS grants a lease with a fixed duration $\delta$ to the localizing FLS.
The localizing FLS must renew this lease prior to its expiration, i.e., $\delta$ time units. 
Otherwise, the anchored FLS will assume the localizing FLS has failed and will un-anchor itself, i.e., change its status to \texttt{available}.

When an anchor FLS fails, as long as the localizing FLS has sufficient information to localize then it localizes, informs the members of its swarm to update their Swarm-ID to that of the failed anchor FLS, and changes its status to \texttt{available}.
SwarMer uses leases to detect failures in an asymmetric manner:
Failure of the anchor does not impact the localizing FLS and leases do not detect a failed anchor FLS.
SwarMer uses leases to detect failure of the localizing FLS, causing the anchor FLS to unanchor itself. 

A standby FLS~\cite{shahram2022} will substitute for a failed FLS as follows.
It uses dead reckoning to move to the location of the failed FLS
and localizes itself relative to a neighbor.

When the battery flight time of an FLS is below a threshold (see STAG~\cite{shahram2022}), it notifies its neighbors and the standby that it is leaving the swarm.
While it flies back to a charging station, a standby FLS assumes its lighting responsibility per the aforementioned discussion of failure handling.

\section{An Evaluation}\label{sec:eval}
This section evaluates SwarMer using several 2D and 3D point clouds, see Figures~\ref{fig:cmp}.a and~\ref{fig:cmp3D}.a.
This evaluation uses a simulation model of SwarMer implemented in MATLAB.
Given a point cloud, it assigns a point to an FLS and implements dead reckoning to compute the location of the FLS deployed from a dispatcher at a well known coordinate.
This dead reckoning technique is described in Section~\ref{sec:dr}.
SwarMer uses this technique repeatedly to localize FLSs.

The simulator executes in rounds. 
At the start of each round, 
an FLS executes the algorithm of Section~\ref{sec:sm}.
We quantify (a) the minimum, average, and maximum number of FLSs that SwarMer localizes in each round, (b) the minimum, average, and maximum distance traveled by FLSs during each round, (c) the number of swarms at the end of each round, (d) the HD at the end of each round.

This evaluation uses standard PNG images for 2D point clouds and the Princeton Shape Benchmark~\cite{princetonbenchmark} for 3D point clouds.
With PNG images, we wrote software to convert it into 2D point clouds, see Figure~\ref{fig:cmp}.a.
The Princeton benchmark consists of a database of 1,814 3D models.
We wrote software to convert its shapes to 3D point clouds.
A few are shown in Figure~\ref{fig:cmp3D}.a.

\subsection{Dead reckoning}\label{sec:dr}
\begin{figure}
\begin{subfigure}[t]{0.47\columnwidth}
\centering
\includegraphics[width=\textwidth]{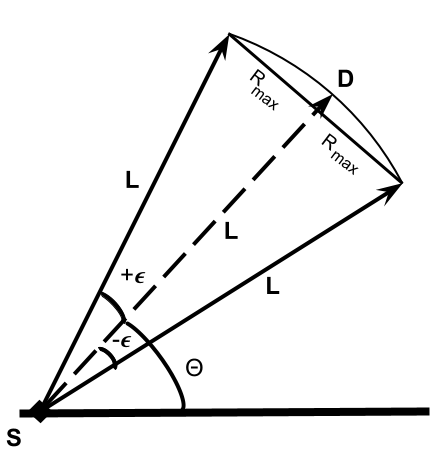}
\caption{2D.}
\label{fig:eval:thpt-stocs}
\end{subfigure}
\quad
\begin{subfigure}[t]{0.45\columnwidth}
\centering
\includegraphics[width=\textwidth]{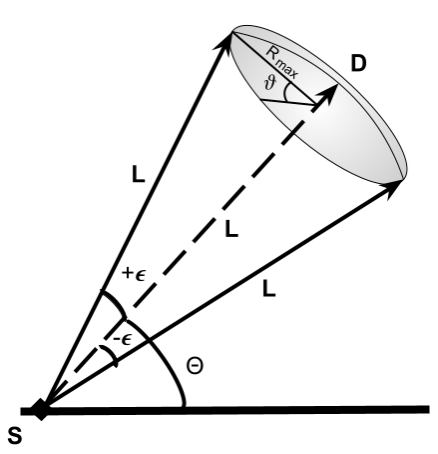}
\caption{3D.}
\label{fig:eval:scale-stocs}
\end{subfigure}
\caption{Angle of error $\epsilon$ with dead reckoning. }
\label{fig:drerr}
\end{figure}

An FLS has a starting position $S$ and a destination $D$ in the ground truth, see Figure~\ref{fig:drerr}.
Vector $\vec{SD}$ at angle $\theta$ has a fixed length $L$.
The error in dead reckoning is bounded by the angle $\epsilon$, $\theta \pm \epsilon$.
In 2D (3D), see Figure~\ref{fig:drerr}.a (\ref{fig:drerr}.b), there is a radius $R_{max}$ 
for the circular-arc (spherical dome) defined by $\epsilon$.
$R_{max}$ increases as a function of $\epsilon$ and $L$.


In 3D, in addition to $\epsilon$, another random number $\vartheta$ between 0 and $2 \pi$ radian (i.e., 0 and 360 degrees) is used to decide the direction in which the epsilon should be added to the vector.
Figure~\ref{fig:drerr}.b shows $\vartheta$ in the spherical dome.

The value of $\epsilon$ is a configuration parameter of the simulator and defines $R_{max}$ in Figure~\ref{fig:drerr}.
Every time an FLS invokes dead reckoning, the simulator generates a random value between $\pm \epsilon$.
It adds this angle to $\theta$ to compute a vector with length $L$ starting at S.  The point $D$ at the end of this vector is the destination of an FLS and its coordinate in the {\em estimated truth}.  The FLS flies to this coordinate.

\subsection{Experimental Results}\label{sec:evalres}

Figure~\ref{fig:cmp} shows SwarMer localizes FLSs to approximate the ground truth with $\epsilon$=5$\degree$, compare Figure~\ref{fig:cmp}.c with Figure~\ref{fig:cmp}.a.
These 2D point clouds consist of approximately 100 FLSs.
SwarMer employs dead reckoning with the Signal Strength technique (see Section~\ref{sec:localize}) and $\eta$=5.
It reduces error by minimizing the distance traveled by localizing FLSs.

\begin{figure}
\centering
\includegraphics[width=3.5in]{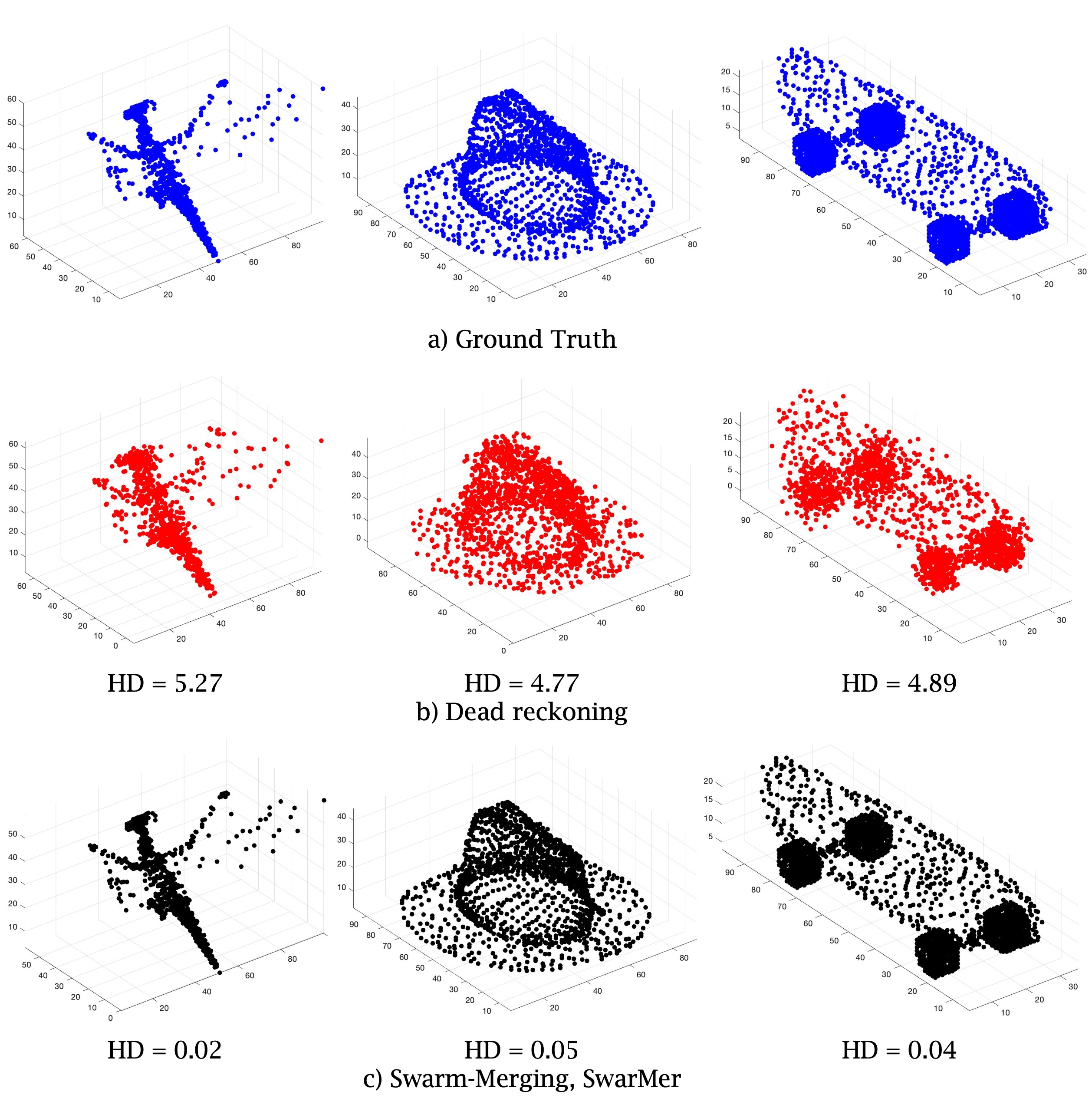}\hfill
\caption{(a) Ground truth for a 3D dragon, a hat, and a skateboard,
(b) illuminations with dead reckoning using $\epsilon$=5$\degree$ and a random $\vartheta$ between 0 and 360 degrees, (c) adjustments with n-ary SwarMer using dead reckoning with $\epsilon$=5$\degree$ and a random $\vartheta$ between 0 and 360 degrees.}
\label{fig:cmp3D}
\end{figure}

Figure~\ref{fig:cmp3D} shows several 3D point clouds from the Princeton Benchmark that challenged our earlier attempts at developing a localizing algorithm.
The dragon, hat, and skateboard\footnote{Their Princeton shape files are m1625, m1630, and m1619 respectively.} consist of 760, 1562 and 1727 points (FLSs), respectively. 
 Figure~\ref{fig:cmp3D}.b shows dead reckoning by itself results in renderings that are distorted.
These are enhanced by SwarMer, resulting in a low HD, see Figure~\ref{fig:cmp3D}.c.
In these experiments, SwarMer merges multiple swarms together at a time, n-ary merging technique of Section~\ref{sec:sm}.
Section~\ref{sec:numswarms} compares this with binary that merges two swarms at a time.

\begin{figure}
\centering
\includegraphics[width=3.65in]{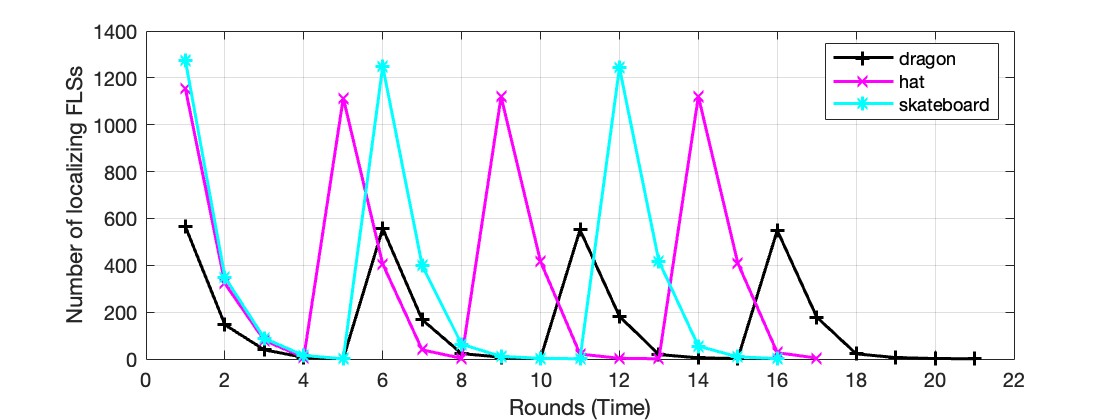}\hfill
\caption{Localizing FLSs for different 3D shapes, $\epsilon$=5$\degree$.}
\label{fig:diff3D}
\end{figure}
Figure~\ref{fig:diff3D} shows the number of localizing FLSs for the different 3D shapes as a function of the number of rounds.
The number is lowest with the dragon because it consists of the fewest number of points.
As the FLSs form swarms, the number of localizing FLSs decreases.
Once there are no localizing FLSs, there is one swarm consisting of all FLSs. 
Different 3D shapes arrive at this point at different rounds:
Round 4 with the hat, and Round 5 with both the dragon and the skateboard.
SwarMer is a continuous framework. 
 Once all FLSs constitute a single swarm, all FLSs thaw their swarm memberships and localize again.
In our simulation studies, we artificially terminated the simulation once there is a single swarm with a HD lower than 0.09.
SwarMer reaches this threshold at a different round for different point clouds.
While it forms and thaws a single swarm twice with the skateboard, the count is thrice with the hat and the dragon; count the peaks in Figure~\ref{fig:diff3D}.


\begin{figure}
\centering
\includegraphics[width=3.65in]{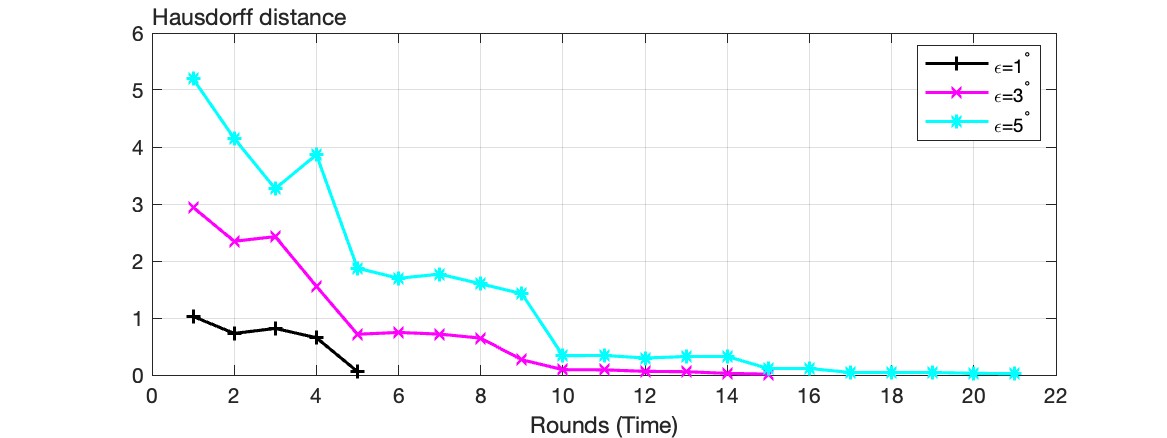}\hfill
\caption{HD.}
\label{fig:hd}
\end{figure}
Below, we focus on the results obtained from the dragon.
Figure~\ref{fig:hd} shows SwarMer's HD with different degrees of error, i.e., $\epsilon$ values of 1, 3, and 5 degrees, as a function of the number of rounds.
A lower HD is better because it shows the rendering approximates the ground truth more accurately.  
This distance decreases as a function of the number of rounds.  
With $\epsilon$=1$\degree$, the simulation terminates in the 5th round because its HD drops below 0.09.

\begin{figure}
\centering
\includegraphics[width=3.65in]{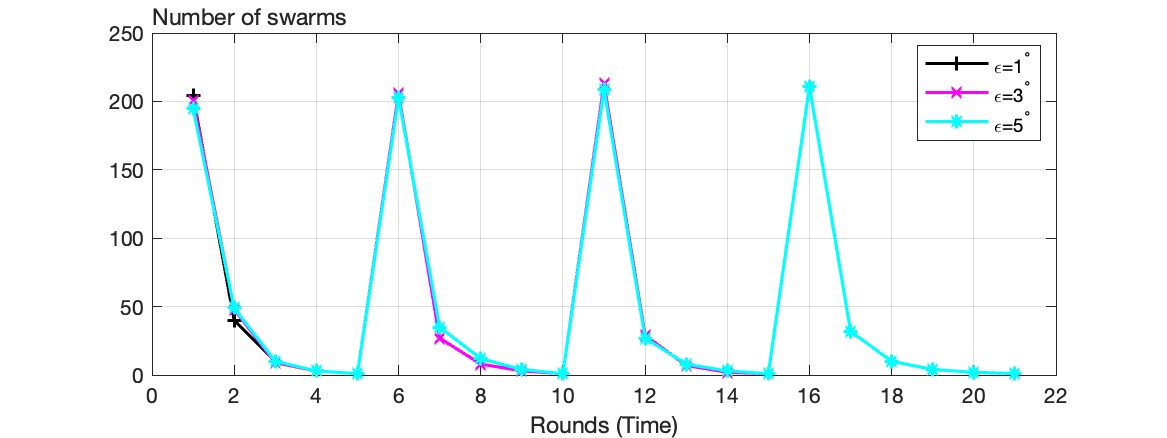}\hfill
\caption{Number of swarms.}
\label{fig:numswarms}
\end{figure}

\begin{figure}
\centering
\includegraphics[width=3.65in]{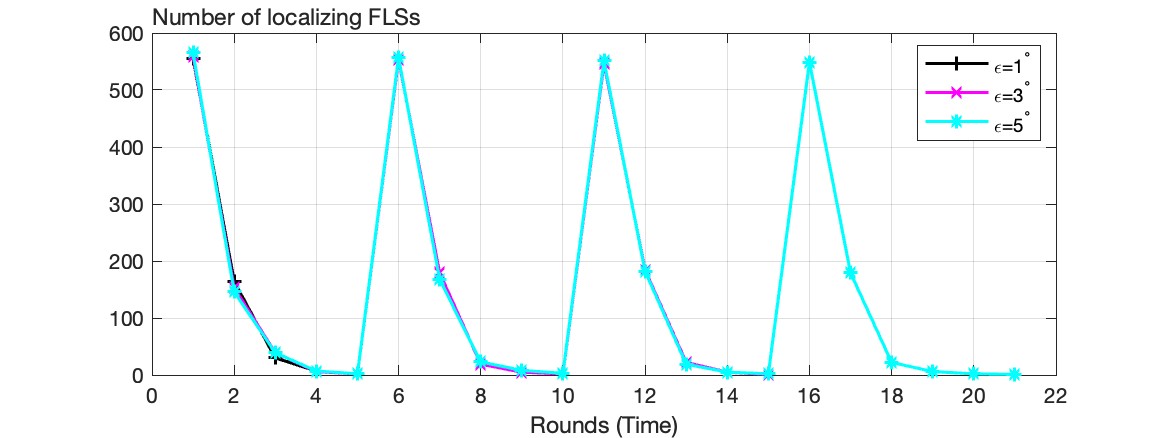}\hfill
\caption{Number of localizing FLSs.}
\label{fig:numslocalizing}
\end{figure}

\begin{figure}
\centering
\includegraphics[width=3.65in]{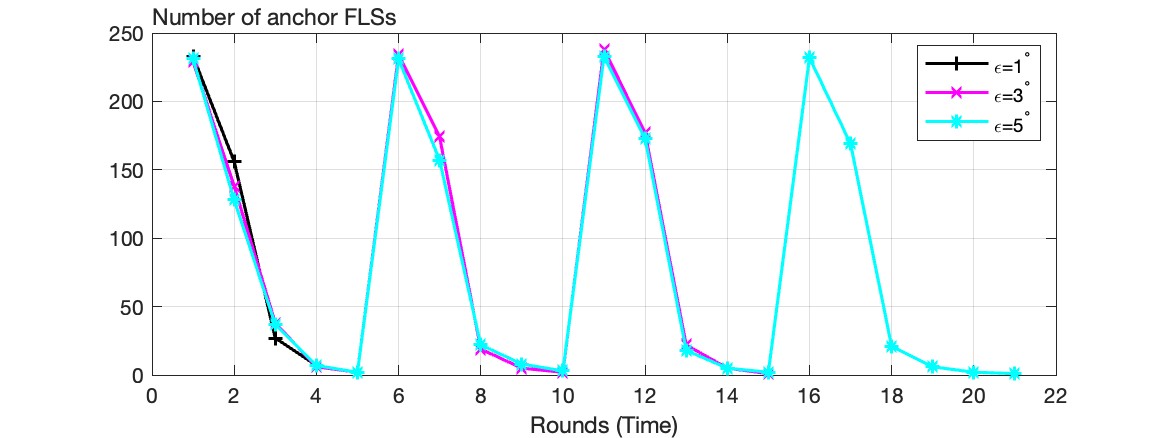}\hfill
\caption{Number of anchor FLSs.}
\label{fig:anchors}
\end{figure}

SwarMer forms swarms starting in the first round, see Figure~\ref{fig:numswarms}.
The number of localizing FLSs in each round is approximately the same for different $\epsilon$ values, see Figure~\ref{fig:numslocalizing}.
It is more than 2x the number of anchor FLSs, see Figure~\ref{fig:anchors} and compare with Figure~\ref{fig:numslocalizing}.
This is because 
multiple swarms are merged in each round, see Figure~\ref{fig:sharedanchors}.

\begin{figure}
\centering
\includegraphics[width=3.65in]{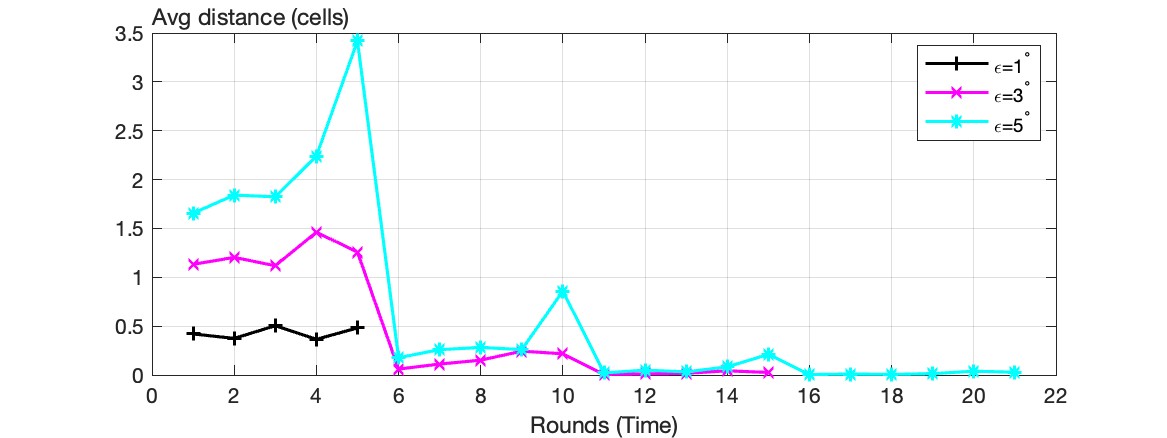}\hfill
\caption{Average dead reckoning distance.}
\label{fig:deadreckoning}
\end{figure}

Figure~\ref{fig:deadreckoning} shows
the average dead reckoning distance as a function of the number of rounds.
This distance is greater with a higher dead reckoning error. 
Generally speaking, this distance decreases as a function of the number of rounds but not always, see $\epsilon$=5$\degree$ in Figure~\ref{fig:deadreckoning}.


\subsubsection{Number of Merged Swarms}\label{sec:numswarms}
An anchor FLS may have multiple localizing FLSs as its candidates.
It may proceed with at least one localizing FLS or multiple of them. 
The former merges two swarms (swarm of the localizing FLS and the anchor FLS) together while the latter may merge a large number of swarms, $\infty$ in theory.
This configuration parameter of SwarMer is denoted as M.
The results presented in the previous section are with M=$\infty$.

When considering the extreme values of M, 2 and $\infty$, M=$\infty$ minimizes the number of rounds to render a shape faster.  
This is true for all shapes considered in our experiments.
For example, it renders the dragon and the skateboard in 21 and 16 rounds, respectively.
M=2 requires 30 and 33 round, respectively.
With the skateboard, M=$\infty$ is more than twice faster when compared with M=2.

\begin{figure}
\centering
\includegraphics[width=3.65in]{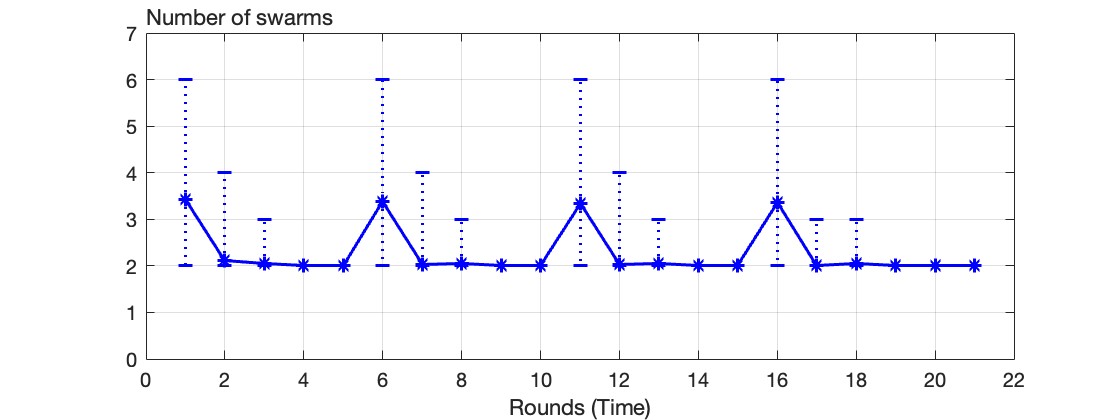}\hfill
\caption{Number of swarms merged in a round, $\epsilon$=5$\degree$, M=$\infty$.}
\label{fig:sharedanchors}
\end{figure}

M=$\infty$ is faster because it merges multiple swarms in each round. 
Figure~\ref{fig:sharedanchors} shows the minimum, average, and maximum number of localizing swarms per anchor FLS for the dragon in different rounds\footnote{Reported results use the technique that selects the FLS belonging to the swarm with the highest number of FLSs as the anchor.}.
The average value is as high as 3.5 swarms merging in a round.  
The maximum is as high as 6 in the first, sixth, eleventh, and sixteenth rounds.
With M=2, the number of merged swarms is always 2 for all rounds.

M=$\infty$ minimizes the overall distance traveled by the FLSs when compared with M=2, compare last two columns of Table~\ref{tbl:distance}.
Traveled distance may be used to quantify the amount of energy consumed by SwarMer to localize FLSs.
Hence, M=$\infty$ may be more energy efficient.
We separate distance traveled by localizing FLSs from the distance traveled by the FLSs that constitute swarms. 
In our experiments, M=2 reduces the distance traveled by localizing FLSs when compared with M=$\infty$, see 2nd and 3rd columns of Table~\ref{tbl:distance}.
However, M=$\infty$ reduces the distance traveled by swarms when compared with M=2.
This distance dominates the total distance, enabling M=$\infty$ to outperform M=2.

\begin{table}
\begin{small}
\begin{center}
\begin{tabular}{||c | c c | c c || c c||} 
 \hline
  & \multicolumn{2}{|c|}{Localizing FLSs} & \multicolumn{2}{|c||}{Swarm} & \multicolumn{2}{|c||}{Total} \\ [0.5ex] 
\cline{2-7}
  & M=2 & M=$\infty$ & M=2 & M=$\infty$ & M=2 & M=$\infty$ \\
 \hline\hline
 Dragon & 1,356 & 1,478  & 7,542 & 4,233 & 8,898 & 5,711 \\ 
 \hline
 Hat & 3,352 & 3,670 & 19,187 & 9,184 & 22,539 & 12,854 \\
 
 \hline
 Skateboard & 2,733 & 2,897 & 17,538 & 10,411 & 20,271 & 13,308 \\
 \hline
\end{tabular}

\end{center}
\end{small}
\caption{Traveled distance (cells)}\label{tbl:distance}
\end{table}



\subsubsection{Techniques to Select Anchor}\label{sec:anchortech}
With M=$\infty$, when an FLS expands its radio range to identify other swarms to merge with, it may encounter multiple FLSs with different Swarm-IDs.
We experimented with the following four techniques to select which FLS should serve as an anchor:  1) Random, 2) the FLS expanding its radio range to discover and challenge other FLSs, 3) the FLS with the lowest Swarm-ID, 4) the FLS belonging to the swarm with the most (fewest) number of FLSs.

We compared the alternative techniques by quantifying the total traveled distance by FLSs.
No one technique is superior across all point clouds.
For a given point cloud, one technique may be superior to another.
For example, with the dragon, the technique that selects the FLS belonging to the swarm with the most FLSs is superior to the rest.
However, with the skateboard, the technique that selects the FLS belonging to the swarm with the fewst FLSs is suprior.
If an FLS renders diverse shapes that are not known in advance then using a simple technique such as the lowest Swarm-ID as the anchor may be sufficient because of its simplicity.  


\subsubsection{Scalability}
\begin{figure}
\centering
\includegraphics[width=3.65in]{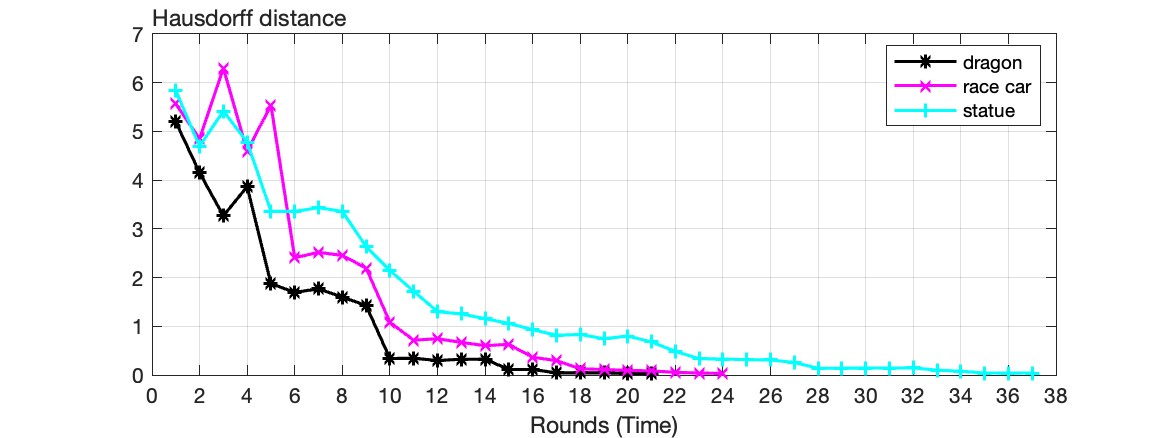}\hfill
\caption{Shapes with tens of thousands of FLSs, $\epsilon$=5$\degree$, M=$\infty$.}
\label{fig:scalehd}
\end{figure}

SwarMer is decentralized and scales to render shapes consisting of many FLSs accurately and quickly.
We evaluated SwarMer using several large point clouds produced using the Princeton shape benchmark~\cite{princetonbenchmark}:
\begin{itemize}
\item m1510, a race car, consisting of 11,834 unique points.  The original consists of 11,894 points with 60 duplicates. 
\item m303, statue of a face, consisting of 22,310 unique points. 
 The original consists of 22,375 points with 65 duplicates. 
\end{itemize}
Figure~\ref{fig:scalehd} shows the number of rounds required for SwarMer to realize a HD lower than 0.09 with $\epsilon$=5$\degree$.
As a comparison, we include the dragon of Figure~\ref{fig:cmp3D} consisting of fewer than a thousand points (760 points).

The number of rounds required to complete the first swarm is 5 with the dragon and 6 with the race car and the statue.
However, the HD is significantly higher with the statue at the sixth round, see Figure~\ref{fig:scalehd}.
While the dragon requires 21 rounds, the race car with more than 10x points requires 24 rounds, and the statue with more than 20x points requires 37 rounds to realize a HD<0.09.

\begin{figure}
\centering
\includegraphics[width=3.5in]{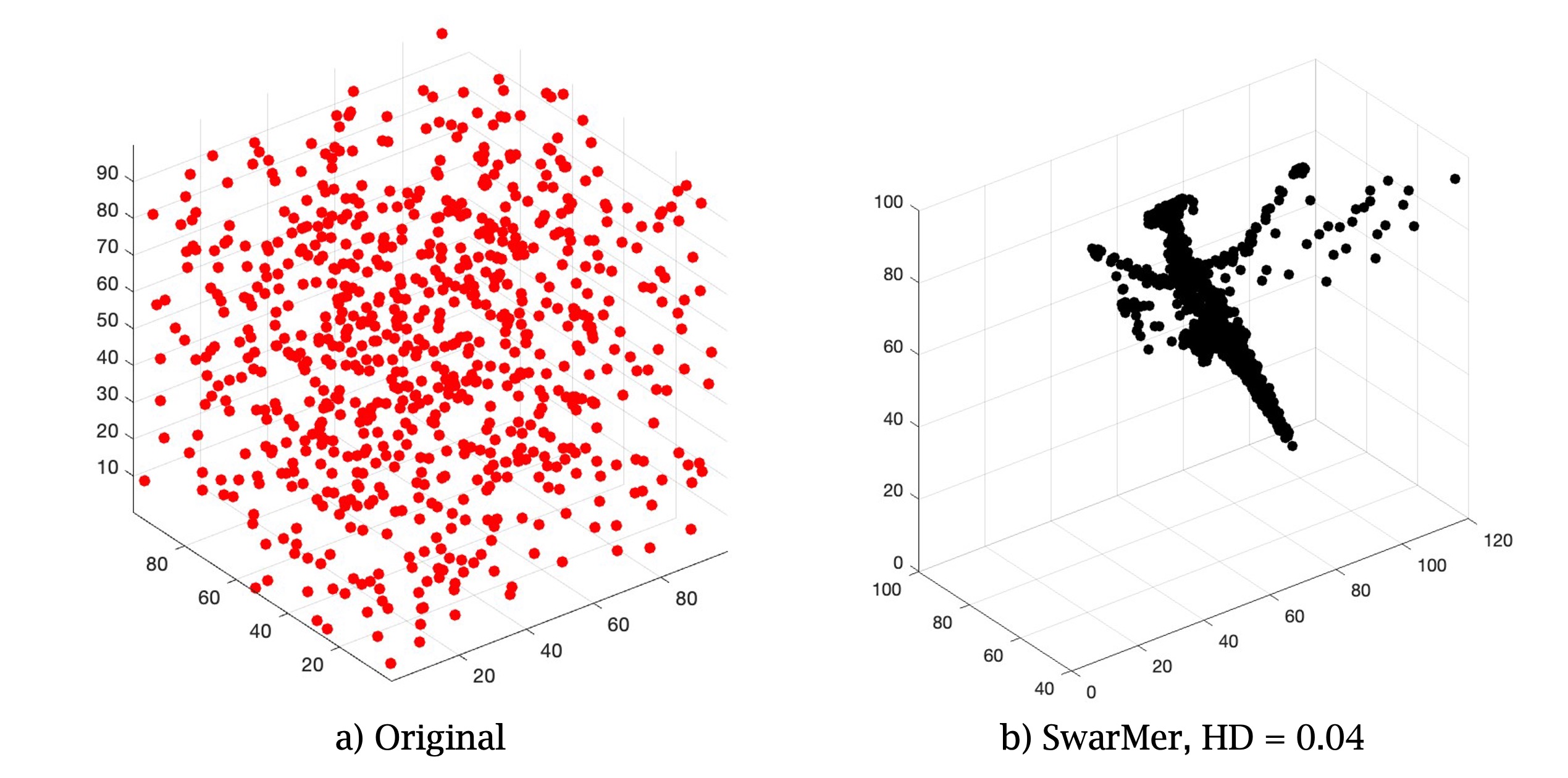}\hfill
\caption{Random placement, see \url{https://youtu.be/cZrz0e61txU}.}
\label{fig:randomPlacement}
\end{figure}

\subsubsection{Random FLS placement}\label{sec:random} 
SwarMer will work with a random placement of FLSs in a 3D display as long as the coordinates of a point are assigned to each FLS.
A limitation is that the shape may be shifted along the different axes by a wide margin because SwarMer is a relative localization technique.
The precise amount of shift depends on how FLSs form swarms, and how swarms merge and move.  
To illustrate, consider the 3D dragon of Figure~\ref{fig:diff3D}.
We assign FLSs to a 3D space using a random number generator, see Figure~\ref{fig:randomPlacement}.a.
FLSs use SwarMer to localize and render the dragon with a Hausdorff distance less than 0.09 in 26 rounds, see Figure~\ref{fig:randomPlacement}.b.
The dragon is shifted along the different axes.
One may address this limitation by having either a reference point on the display with a well known position or an {\em oracle} FLS configured with specialized hardware that computes its position with a high accuracy.
A swarm that localizes relative to either inherits the oracle property.
All other swarms are required to localize relative to its FLSs, preventing a shape from shifting across different axes.

\section{A Comparison}\label{sec:cmp}
In our first approach to localize FLSs to render shapes, we considered the feasibility of using triangulation and trilateration.
We hypothesized that using those FLSs with a high confidence in their location as anchors will produce shapes with a high accuracy.
This hypothesis proved wrong because trilateration and triangulation compound small errors in anchor locations to produce distorted shapes.  
Below, we describe how an FLS computes its confidence.
Subsequently, we describe triangulation and trilateration in turn.
This section ends with a comparison of these techniques with SwarMer.

\noindent{\bf Confidence:}  Once an FLS $i$ ($F_i$) arrives at its destination, it may compute its confidence $C_i$ using either the average or worst estimated error attributed to dead-reckoning.  
With average (worst), it computes the average (maximum) radius R around the semi-circle in 2D (semi-sphere in 3D) as the error in its position relative to each of its $n_i$ neighbors.
Its confidence is defined as:
\begin{equation}\label{eq:confidence}
    C_i = 1 - \sum_{k=1}^{n_i}{
min(\frac{1}{n_i}, \frac{
R
}{\delta_{gt}(F_i,F_k)})
}
\end{equation}
where $\delta_{gt}(F_i,F_k)$ is the distance between FLS $i$ and a neighboring FLS $k$ in the ground truth.  When a neighbor is missing, $R$ is set to a large value, e.g., max integer.  This causes Equation~\ref{eq:confidence} to use $\frac{1}{n_i}$ for the missing neighbor.  A large value of $C_i$ means the FLS's estimated truth reflects the ground truth more accurately.
$C_i$ equal to 1.0 is ideal.

\begin{figure}
\centering
\includegraphics[width=3in]{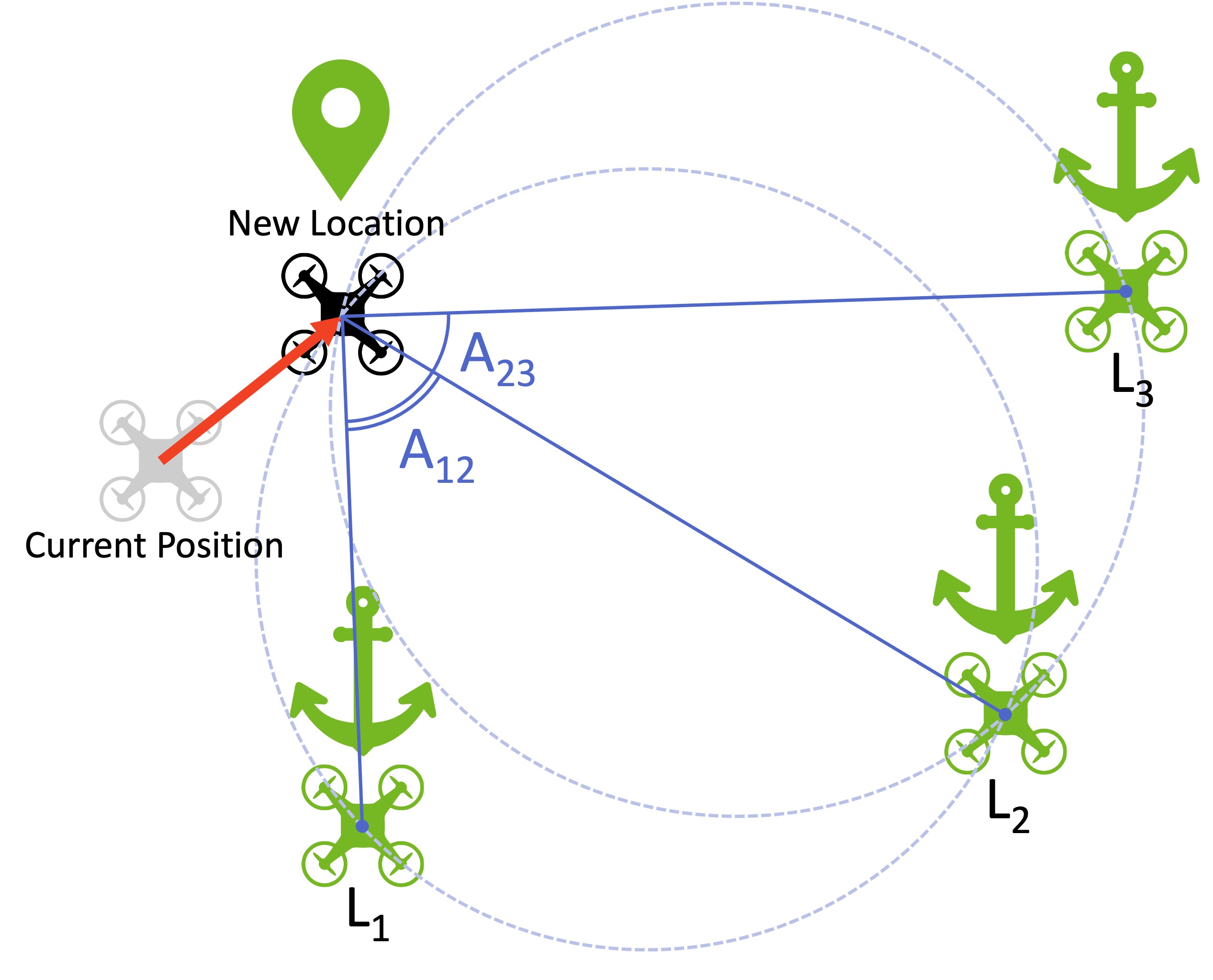}\hfill
\caption{Triangulation}
\label{fig:triangulation}
\end{figure}

\noindent{\bf Three Object Triangulation} requires a localizing object to compute its location and orientation using its angles relative to three objects.
Alternative techniques include iterative search, geometric circle intersection, and Newton-Raphson iterative method~\cite{cohen1993}.
We use the geometric circle intersection method since it is widely used in the literature and ranked as the best when compared with the other two techniques~\cite{cohen1993}.

The geometric circle intersection method is fast and its failures are easy to detect.
It requires a localizing FLS, the one with the lowest confidence relative to its neighbors, to compute two circles using the estimated location and the ground truth angle of three FLSs.
See Algorithm~\ref{alg:triangulate} and Figure~\ref{fig:triangulation}.
The three FLSs may be its neighboring FLSs.  
In most cases, the two circles intersect at two points.  These points are the mutual FLSs and the location of the localizing FLS.  
The latter has an angle relative to the other three FLSs in the ground-truth.  
If these angles match those in the estimated truth then the localizing FLS does not move.  Otherwise, the localizing FLS computes a new location for itself that matches the angles in the ground truth.
The localizing FLS flies to this new location.


\begin{algorithm}
\caption{Triangulation, Circle Intersection Method}\label{alg:triangulate}
\begin{small}
  $f \gets FLS~with~the~lowest~confidence~in~its~location$\;
  $\{L\}= 3~randomly~selected~neighbors~of~f$ \Comment*[r]{Return error if F has fewer than 3 neighbors}
  $\{A_{12}, A_{23}\} = angles~of~f~to~\{L\}~in~the~ground~truth, Figure~\ref{fig:triangulation}$\;
  $Compute~estimated~location~of~\{L\}~in~the~estimate~truth$\;
  $Compute~a~new~location~for~f~such~that~its~angle~to~\{L\}~in~$\newline$the~estimated~truth~is~the~same~as~the~ground~truth$\;
  $Move~f~to~the~location~computed~in~Step~5$\;
  $Repeat~the~above~6~steps~until~each~FLS’s~confidence~exceeds$\newline$~a~prespecified~threshold$
\end{small}
\end{algorithm}

\begin{figure}
\centering
\includegraphics[width=3in]{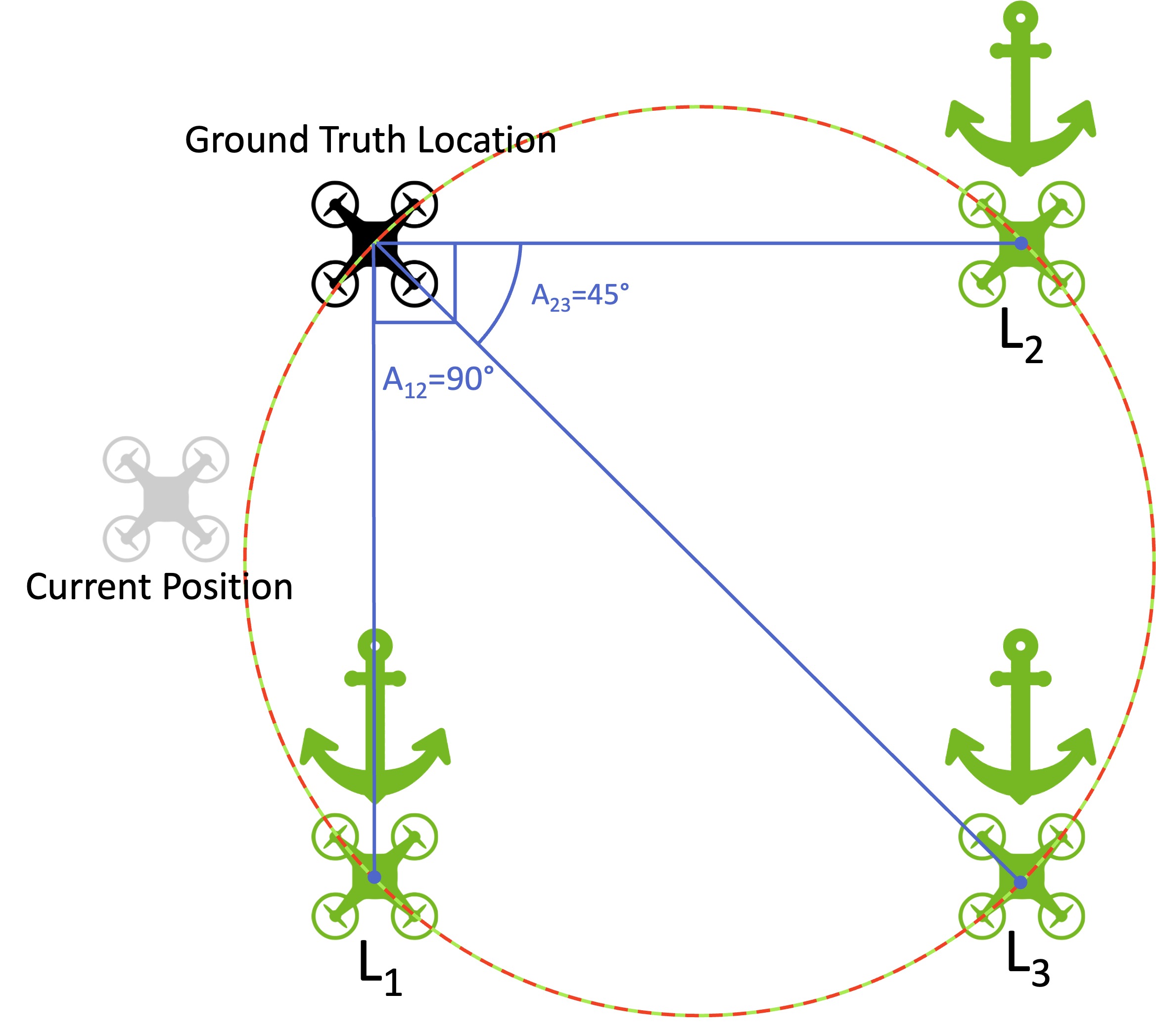}\hfill
\caption{Failure of Triangulation.}
\label{fig:triangulationfailure}
\end{figure}

This method fails when the center of the two circles is too close (we used 1 illumination cell) or when two circles overlap, see Figure~\ref{fig:triangulationfailure}.  


\begin{figure}
\centering
\includegraphics[width=3in]{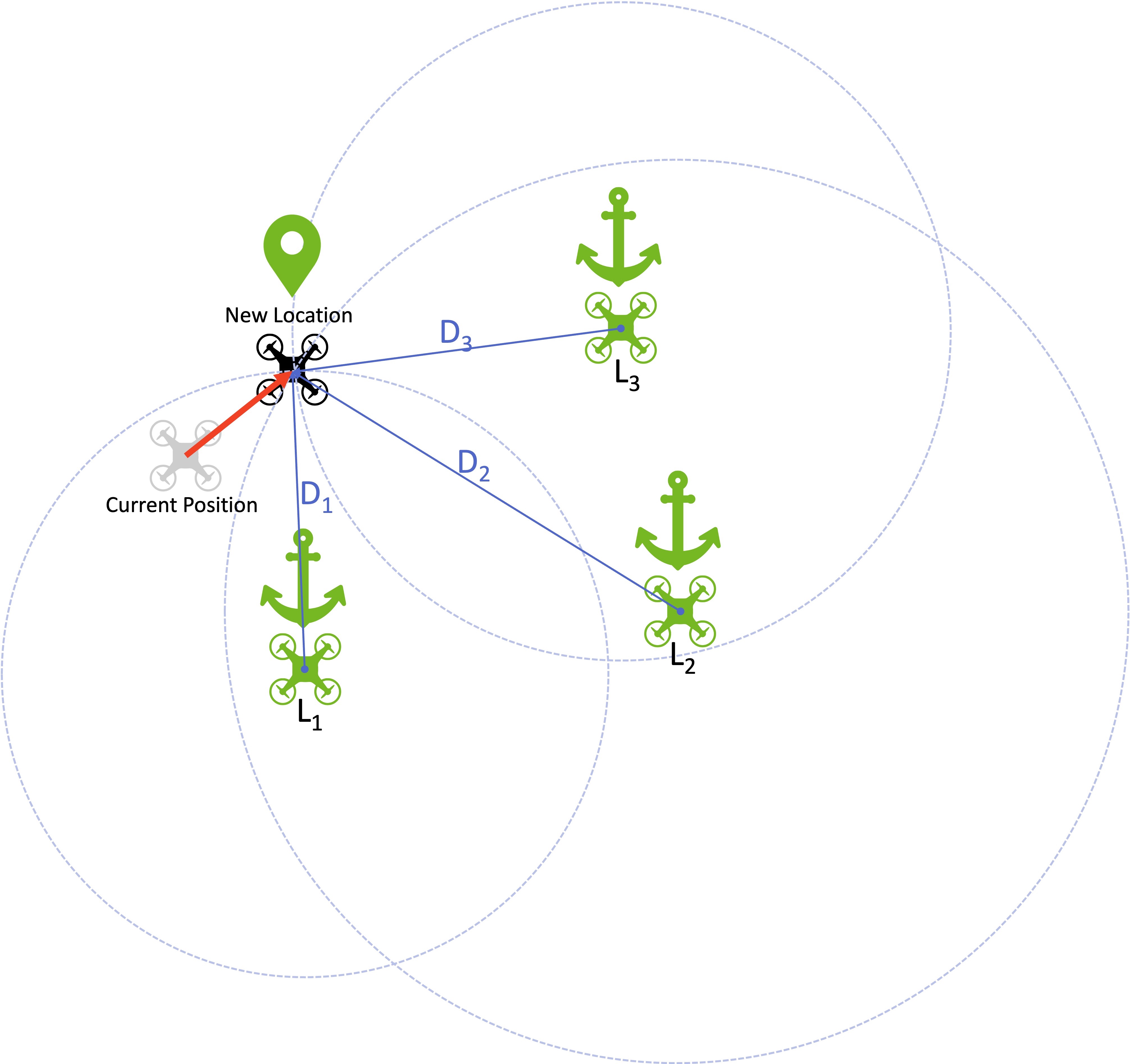}\hfill
\caption{Trilateration}
\label{fig:trilateration}
\end{figure}

\noindent{\bf Three Object Trilateration} is the use of distances or ranges for determining the unknown location of an object.  The localizing FLS $i$ computes its distance to three neighboring FLSs in the ground truth: 
$D_1$, $D_2$, and $D_3$.
See Figure~\ref{fig:trilateration}.
Each delta is the radius of a circle with the corresponding neighbor as its center.
FLS $i$ computes a new estimated location that matches $D_1$, $D_2$, and $D_3$ in its estimated truth.
This is the new location of FLS $i$.

Trilateration fails when either the three circles do not intersect at the same point in the ground truth, or there is no location in estimated truth that matches $D_1$, $D_2$, and $D_3$.
While rare, both do occur in our experiments.

Our implementation of both triangulation and trilateration 
assumes Signal Strength (SS) measurements in the estimated truth match those of the ground truth with 100\% accuracy.
It also assumes an FLS is able to measure angle with triangulation and distances with trilateration 
with 100\% accuracy.
We made these assumptions because triangulation and trilateration perform poorly.
We wanted to remove error in measuring angle and distance as a possible explanation for their inferior localization.

\begin{algorithm}
\caption{Trilateration}\label{alg:trilateration}
\begin{small}
  $f \gets FLS~with~the~lowest~confidence~in~its~location$\;
  $\{L\} = 3~randomly~selected~neighbors~of~f$ \Comment*[r]{Return error if F has fewer than 3 neighbors}
  $\{D\} = distance~of~f~to~\{L\}~in~the~ground~truth, Figure~\ref{fig:trilateration}$\;
  $Compute~location~of~\{L\}~in~the~estimated~truth$\;
  $Compute~a~new~location~for~f~such~that~its~distance~to~\{L\}~in~$\newline$the~estimated~truth~is~the~same~as~the~ground~truth$\;
  $Move~f~to~the~location~computed~in~Step~5$\;
  $Repeat~the~above~6~steps~until~each~FLS’s~confidence~exceeds$\newline$~a~prespecified~threshold$
\end{small}
\end{algorithm}

\subsection{Experimental Results}

\begin{figure}
\centering
\includegraphics[width=3.4in]{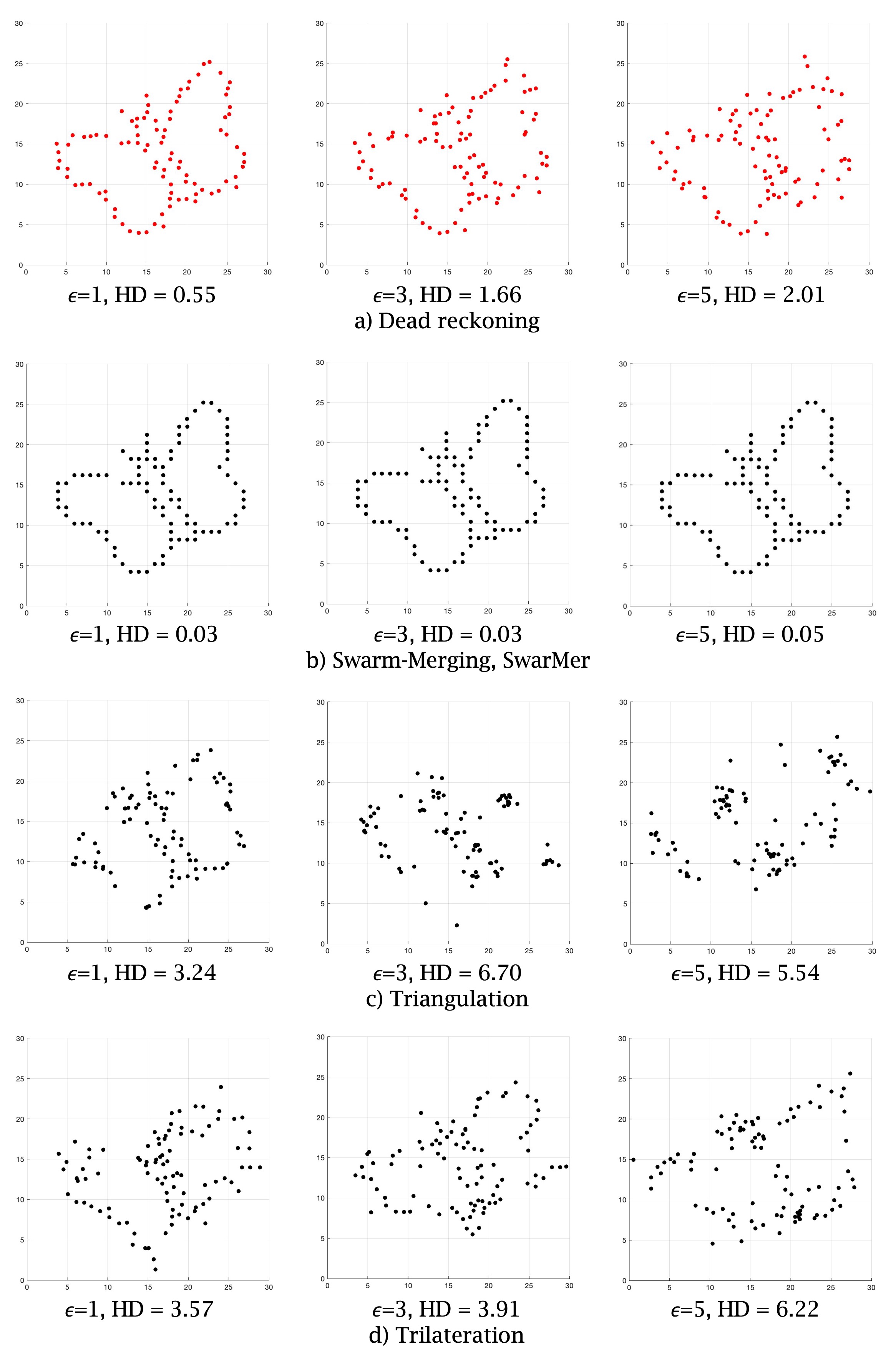}\hfill
\caption{(a) A 2D butterfly with dead reckoning and three different degrees of error,
(b) with SwarMer, (c) with Triangulation, and (d) Trilateration.}
\label{fig:cmpbutterfly}
\end{figure}

Figure~\ref{fig:cmpbutterfly}.a-d show rendering of a 2D butterfly using dead reckoning with different degrees of error, $\epsilon$=$\{1\degree,3\degree,5\degree\}$.
Figure~\ref{fig:cmpbutterfly}.b shows application of SwarMer to each butterfly.
It continues to use the same $\epsilon$ when applying dead reckoning using signal strength.
The last two rows of Figure~\ref{fig:cmpbutterfly} show triangulation and trilateration do not improve the quality of illumination.
With both, an FLS uses the worst estimated error attributed to dead-reckoning to compute its confidence.
Trilateration decreases the quality of rendering when compared with dead reckoning with $\epsilon$=1$\degree$.
This is despite the fact that both triangulation and trilateration are provided with 100\% accurate angle and distance measurements using signal strength.
Similar results are observed with the cat and the teapot in Figure~\ref{fig:cmp}.
The key difference is that Figure~\ref{fig:cmp} reports results with $\epsilon$=5$\degree$ while Figure~\ref{fig:cmp} shows different degrees of error, $\epsilon$=$\{1\degree,3\degree,5\degree\}$.

\begin{figure}
\centering
\includegraphics[width=3.4in]{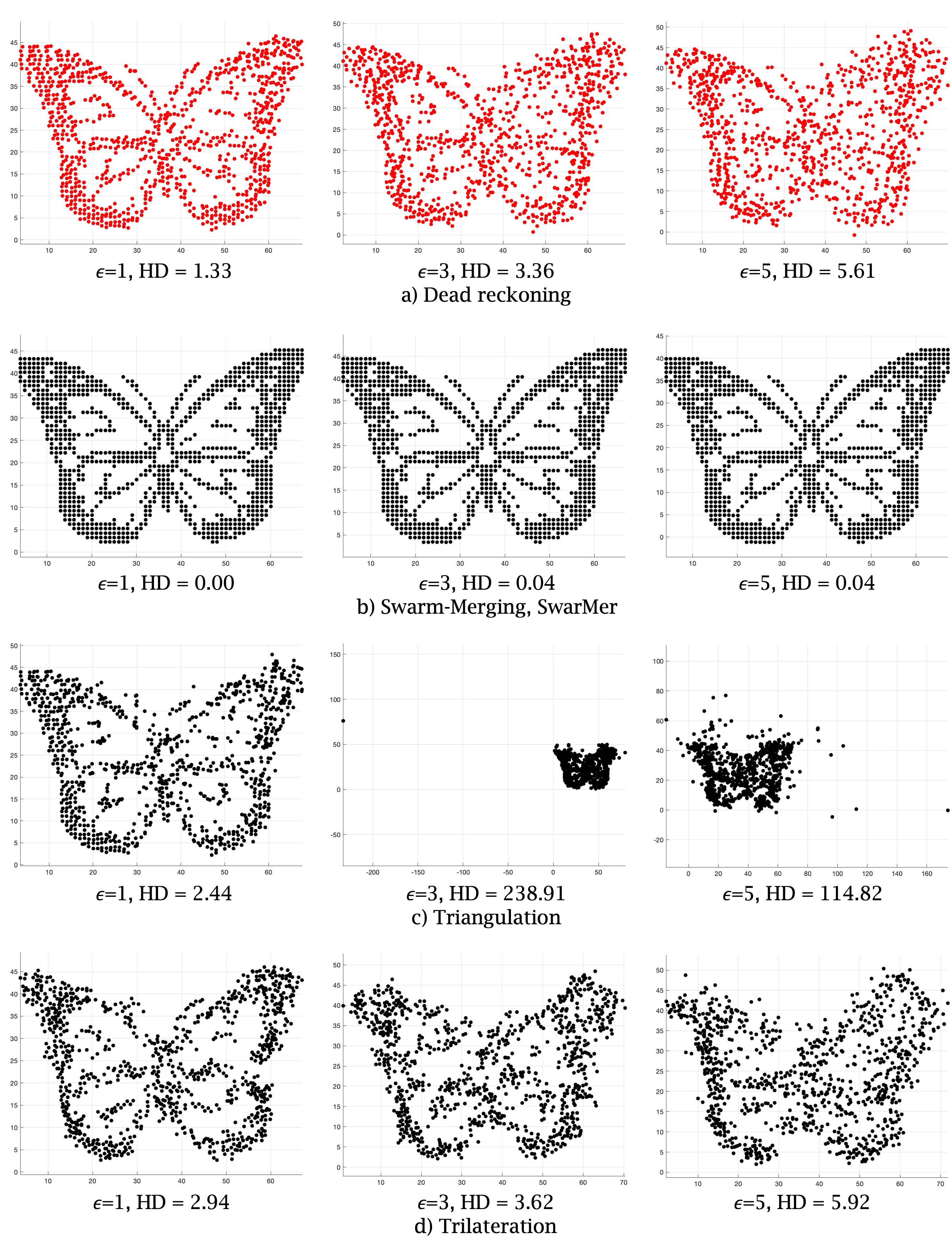}\hfill
\caption{(a) A 2D butterfly with dead reckoning and three different degrees of error,
(b) with SwarMer, (c) with Triangulation, and (d) Trilateration..}
\label{fig:bigbutterfly}
\end{figure}

The number of points in a point cloud impacts the quality of renderings with small values of $\epsilon$.
Figure~\ref{fig:bigbutterfly} shows a 2D butterfly consisting of 1008 points, more than ten times the number of points in the butterfuly of Figure~\ref{fig:cmpbutterfly}.
The FLSs deployed from the origin, (0,0), travel the longest distance to the rightmost wing of the butterfly, incurring the highest amount of error and the most distortion with $\epsilon$=5$\degree$.
With $\epsilon$=1$\degree$, the shape of the butterfly is recognizable with both triangulation and trilateration.
However, their Hausdorff distance is worse than dead reckoning by itself.
With $\epsilon$=3 and 5 degrees, triangulation results in a few FLSs to move away from the illumination by a large distance.
This is reflected in their longer scale of the x-axis in Figures~\ref{fig:bigbutterfly}c.
While trilateration does not suffer from this limitation, its localization results in distorted shapes with a large Hausdorff distance.
SwarMer continues to provide accurate localization subjectively, see Figures~\ref{fig:bigbutterfly}b.  
It realizes a Hausdorff distance lower than 0.09 by executing 20, 30, and 40 rounds with $\epsilon$=1, 3, and 5 degrees, respectively.

\section{An Implementation}\label{sec:impl}
We implemented SwarMer as a process to execute on board an FLS.
Processes exchange messages to implement SwarMer.
Our current implementation uses the Python 3.9 programming language.
We used a cluster of Amazon AWS instance \href{https://aws.amazon.com/ec2/instance-types/high-memory/}{u-6tb1.112xlarge} to conduct the experiments reported in this section.
Each instance consists of 448 virtual cores and 6 Terabytes of memory.
While the Dragon requires a cluster of 6 such servers, the Hat and the Skateboard use a cluster of 16 such servers because they consist of more points (FLSs).

Each process is termed $P_{FLS}$ and implements a finite state machine that maintains the following 5 states:
Deploying, Available, Busy Anchor, Busy Localizing, and Waiting.
11 events corresponding to different message types transition the state of a $P_{FLS}$ to implement SwarMer.

A $P_{FLS}$ consists of two threads:
A networking thread ($T_{Network}$) and a handler thread ($T_{Handler}$) that mutates the state of the machine.
While $T_{Network}$ only reads the FLS state,
$T_{Handler}$ may read and write the FLS state.
$T_{Network}$ blocks on the network socket and processes messages as fast as they arrive by reading the current state of the machine.
One of its objectives is to release the lease granted to an FLS that has decided to become an anchor FLS itself\footnote{This happens due to an undesirable race condition.  An example race condition is FLS 5 challenging FLSs 4 and 6.  FLS 6 accepts the challenge, causing FLS 5 to become its anchor.  Concurrently, FLS 4 accepts FLS 5's challenge to become its anchor by granting a lease to FLS 5.  Should FLS 5 enter the waiting state for FLS 6, its $T_{Network}$ reads Waiting state, detects the undesirable race condition, and informs FLS 4 to unanchor itself and release its lease.  This minimizes the number of lease expirations, expediting SwarMer.}.  In addition, it is designed to
minimize latency between cooperating FLSs, expediting the execution of SwarMer to form shapes faster.
$T_{Network}$ places those messages that change system state in a queue for processing by $T_{Handler}$.
$T_{Handler}$ consumes these messages from the queue and may transition and update the state of the machine.
The queue introduces delays and $T_{Network}$'s processing of certain messages minimizes the impact of this delay on a swarm of cooperating FLSs.

This implementation is different from the Matlab simulation model in several ways.  First, each $P_{FLS}$ implements the concept of time by either sleeping or busy-waiting for the duration of time it travels from its current location to a new destination.  It has a realistic velocity model that controls its acceleration and deceleration to arrive at its destination with a speed of zero.  Second, each $P_{FLS}$ implements a UDP-Wrapper that emulates a networking card with an adjustable radio range.  The time required for a $P_{FLS}$ to expand its radio range and transmit messages is another difference from the Matlab simulator.
Third, a $P_{FLS}$ may incur different forms of failures such as packet loss, and a process shutdown to simulate FLS failures.
Finally, a $P_{FLS}$ uses a probabilistic method to thaw the final swarm.
With this method, a $P_{FLS}$ sets a timer to expire between $H$ and $2H$ seconds where $H=log_2F$ and F is the number of FLSs.
Once an FLS's timer expires, it broadcasts a thaw message using its maximum radio range.
A receiving FLS sets its Swarm-ID to its FID and resets its timer.
 
The lessons learned from the emulator are as follows:
\begin{itemize}
\item Within a few seconds, SwarMer reduces the HD by several orders of magnitude.
\item Packet loss as high as 10\% does not hinder SwarMer's ability to reduce the HD in the first few seconds. 
It is able to construct a single swarm in the presence of packet loss.  
\item Today's small drones may travel at speeds as fast as 3 to 5 meters/second.  While FLS speed impacts the initial deployment of FLSs using dead reckoning, it does not impact 
how fast SwarMer improves the quality of an illumination (i.e., drop in HD) as a function of time.
The key insight is that the average distance moved by an FLS (or swarm) to localize is small.
This prevents an FLS from accelerating to its maximum speed.
\item SwarMer is resilient to FLS failures.  It incorporates FLSs that replace failed FLSs seamlessly.  
\item FLS latency in processing messages may result in undesirable race conditions that slow-down SwarMer without impacting its functionality.
We present a multi-threaded $P_{FLS}$ to minimize processing latency.
\end{itemize}

In our experiments with drones, we have observed the communication link to drop packets.  
A $P_{FLS}$ uses UDP (instead of TCP) to communicate with other $P_{FLS}$s.  UDP is an unreliable communication protocol.  It may drop packets, deliver a sequence of packets transmitted by a $P_{FLS}$ out of order, and delay the delivery of packets.  Our implementation of SwarMer is non-blocking to accommodate the first two features of UDP.  With UDP's last two features, we introduced a monotonically increasing message id for each packet sent by a $P_{FLS}$.  A $P_{FLS_j}$ maintains the largest message id it has received from another $P_{FLS}$ and may drop those with a smaller id.  

A transmitting $P_{FLS}$ uses UDP in broadcast mode.
The emulated processes run on one or more servers.  A message sent by one process is received by all processes.  We built a thin networking layer around UDP, a UDP-Wrapper, to implement an adjustable radio range.  The UDP-Wrapper implements a networking card with a fixed radio range.
It uses the distance between FLSs in the ground truth to decide whether a $P_{FLS}$ receives a UDP message broadcasted by another $P_{FLS}$.
When a process (i.e., an FLS) broadcasts a message within a fixed radio range, only the processes (FLSs) in this radio range receive the message.  

Pushing features into the UDP-Wrapper expedites the processes on a multi-core server.  For example, when a member of the swarm issues the challenge message, all FLS processes in the same swarm receive the message due to the broadcast nature of UDP.  The UDP-Wrapper may compare the Swarm-ID of a process with the Swarm-ID of the challenger (embedded in the message), dropping the packet with a matching Swarm-ID\footnote{Members of the same swarm may not challenge one another.}.  Such optimizations enhance performance and reduce the variation in the duration of time required to form a single swarm.

\begin{table*}
\begin{small}
\begin{center}
\begin{tabular}{|| c ||c  c  c  c  c  c ||} 
 \hline
 \hline

 $\lambda$ & Total Xmit bytes & \# Localizations &  \# Anchor & \# Thawed swarms & \# Leases expired & Avg distance traveled \\

\hline
\hline

50 ms & 706,152,707 & 41,814 & 39,569 & 1,507,271 & 8606 (11.03\%) & 70.38 \\
\hline
200 ms & 284,105,583 & 15,218 & 16,131 & 468,330 & 2468 (6.92\%) & 68.37 \\
\hline
300 ms & 381,641,949 & 21,050 & 22,279 & 830,543 & 3351 (6.78\%) & 69.92 \\
\hline
450 ms & 235,055,229 & 12,582 & 13,511 & 553,477 & 2013 (6.53\%) & 72.68 \\
\hline
700 ms & 153,765,307 & 8,406 & 9,149 & 335,810 & 1158 (5.35\%) & 74.67 \\
\hline
1000 ms & 102,582,418 & 5,427 & 5,892 & 170,807 & 746 (5.16\%) & 75.24 \\
\hline
\end{tabular}

\end{center}
\end{small}
\caption{System metrics as a function of $\lambda$ for a two minute execution of SwarMer.}\label{tbl:ci_metrics}
\end{table*}

\begin{figure}
\centering
\includegraphics[width=\columnwidth]{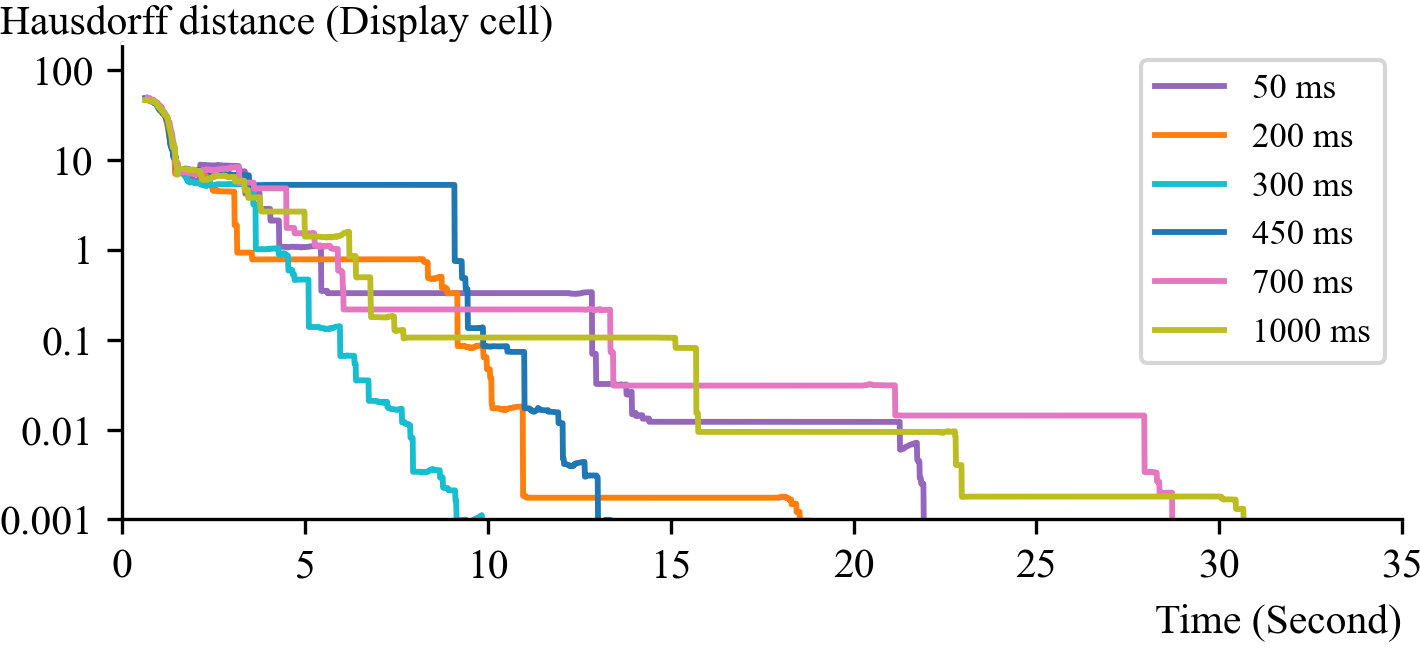}
\caption{HD with different challenge inter-arrival times, $\lambda$ for a 3D figure sampled with 200 points from the Chess piece.  Video clips showing the illuminations with $\lambda$ values 50, 450, and 1500 msec intervals available at \url{https://youtu.be/zwTUgKvg6sI}, \url{https://youtu.be/AKJFu05eM1Y}, and \url{https://youtu.be/8_OtMU4bE74}, respectively.}
\label{fig:HDimpl}
\end{figure}

\begin{figure}
\centering
\includegraphics[width=\columnwidth]{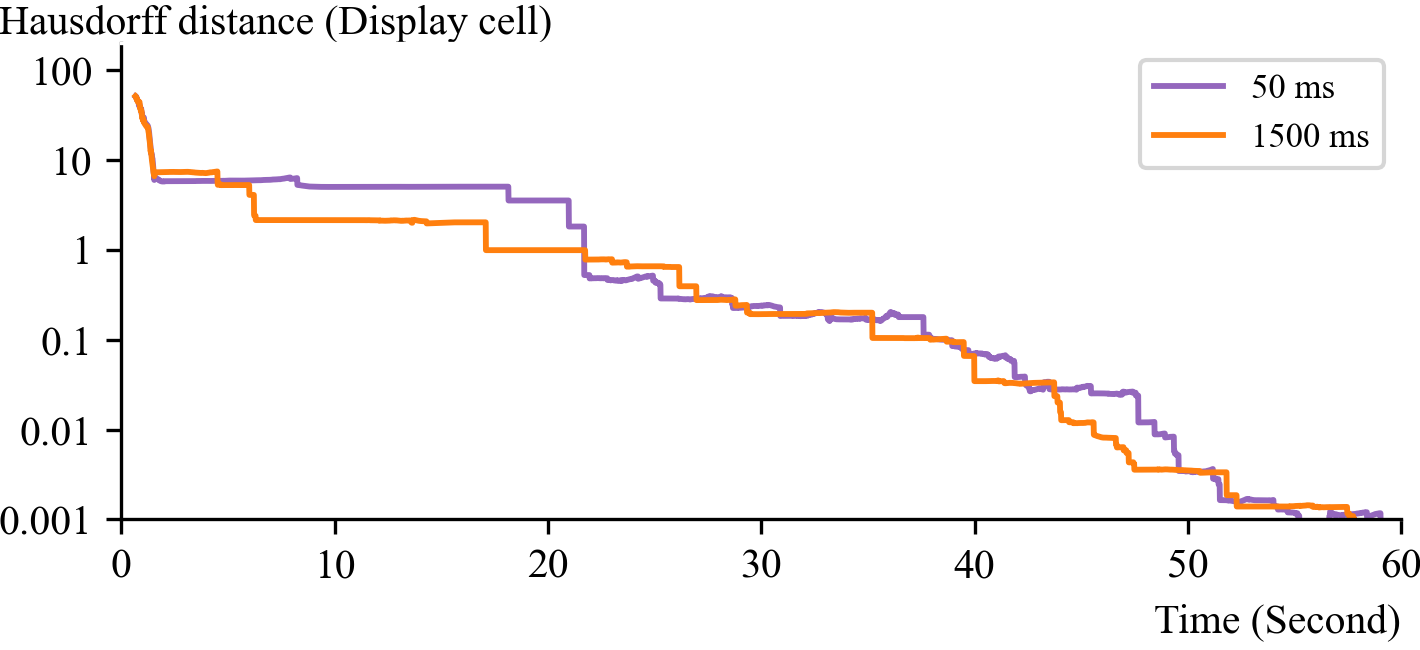}
\caption{HD with different challenge inter-arrival times, $\lambda$ for the Skateboard.  Video clips with $\lambda$ values 50 and 450 available at \url{https://youtu.be/WdjsTJp1RNY} and \url{https://youtu.be/pzaOrDuvL3g}, respectively.}
\label{fig:HDimpl}
\end{figure}

A key parameter of SwarMer is the inter-arrival time $\lambda$ between the Challenge messages issued by $T_{Handler}$ thread of an FLS to its neighbors.
Figure~\ref{fig:HDimpl} shows the HD with the different values of $\lambda$.
Table~\ref{tbl:ci_metrics} shows key system metrics for different $\lambda$ values.
Experiments below use $\lambda$=500 msec.

\begin{figure*}[t!]
    \centering
    \begin{subfigure}[b]{0.33\textwidth}
        \centering
        \includegraphics[width=\textwidth]{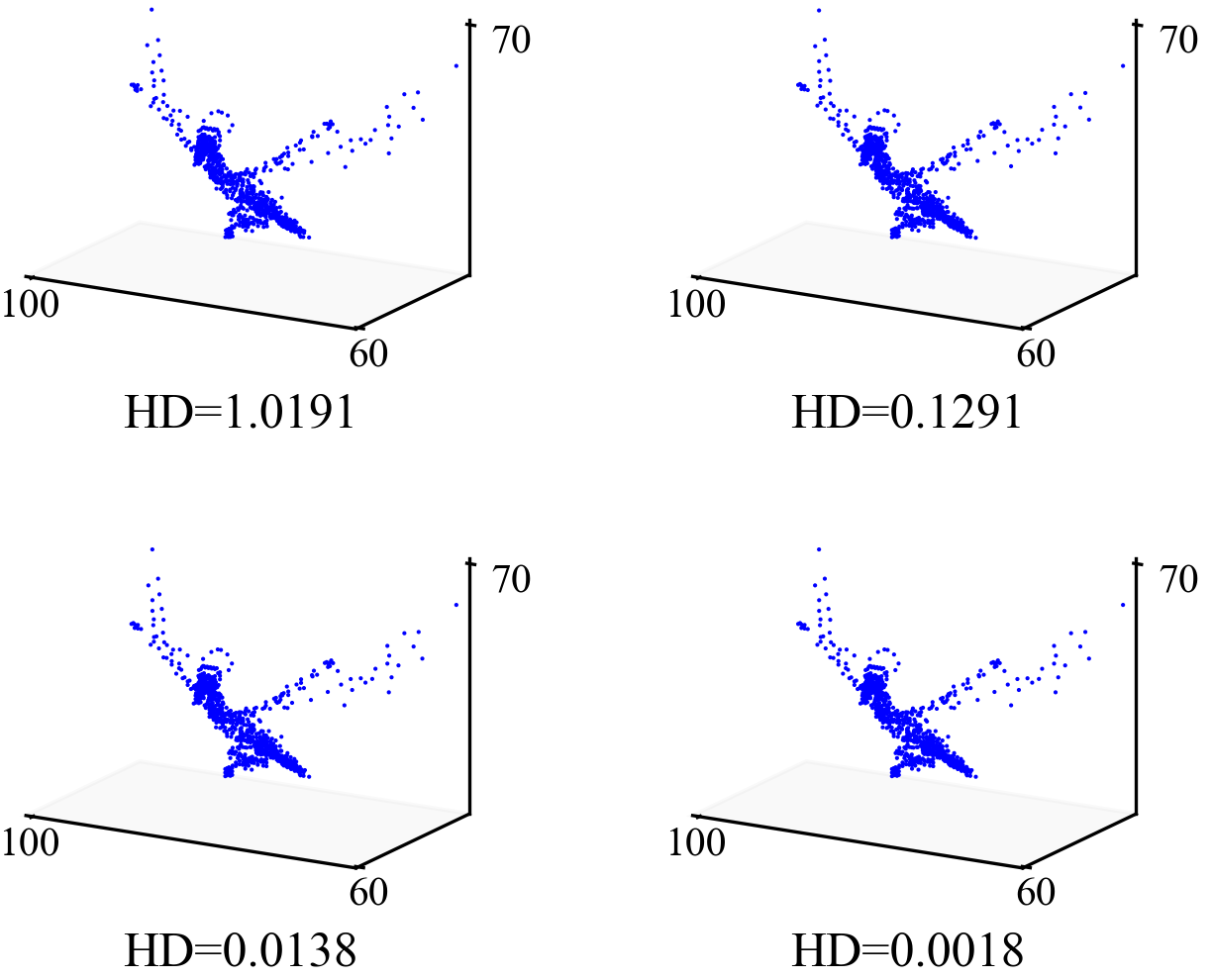}
        \caption{Dragon, \url{https://youtu.be/PRyjdmw8NVc}}
    \end{subfigure}%
    ~ 
    \begin{subfigure}[b]{0.33\textwidth}
        \centering
        \includegraphics[width=\textwidth]{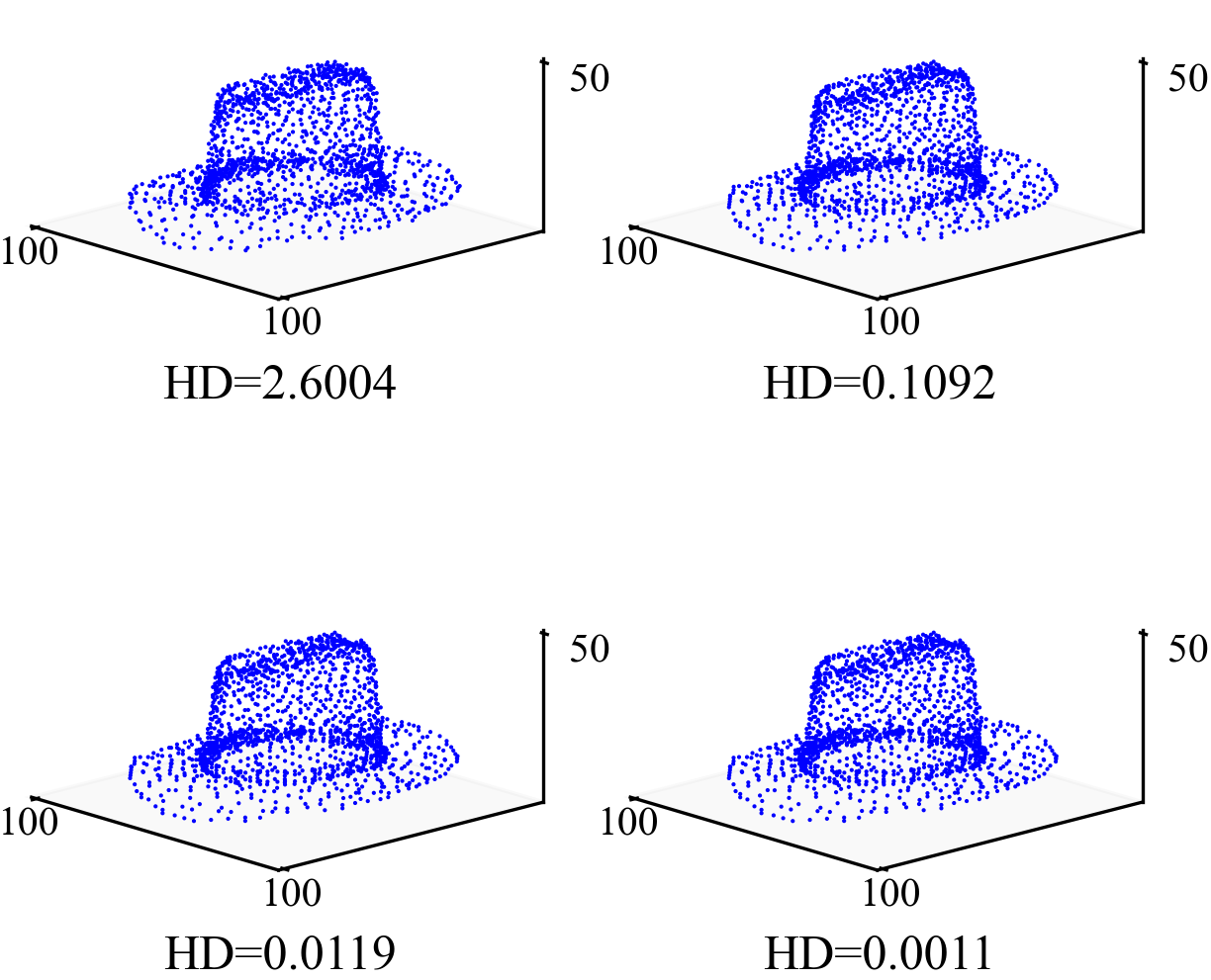}
        \caption{Hat, \url{https://youtu.be/y_xHdz5bs5M}}
    \end{subfigure}
    ~ 
    \begin{subfigure}[b]{0.33\textwidth}
        \centering
        \includegraphics[width=\textwidth]{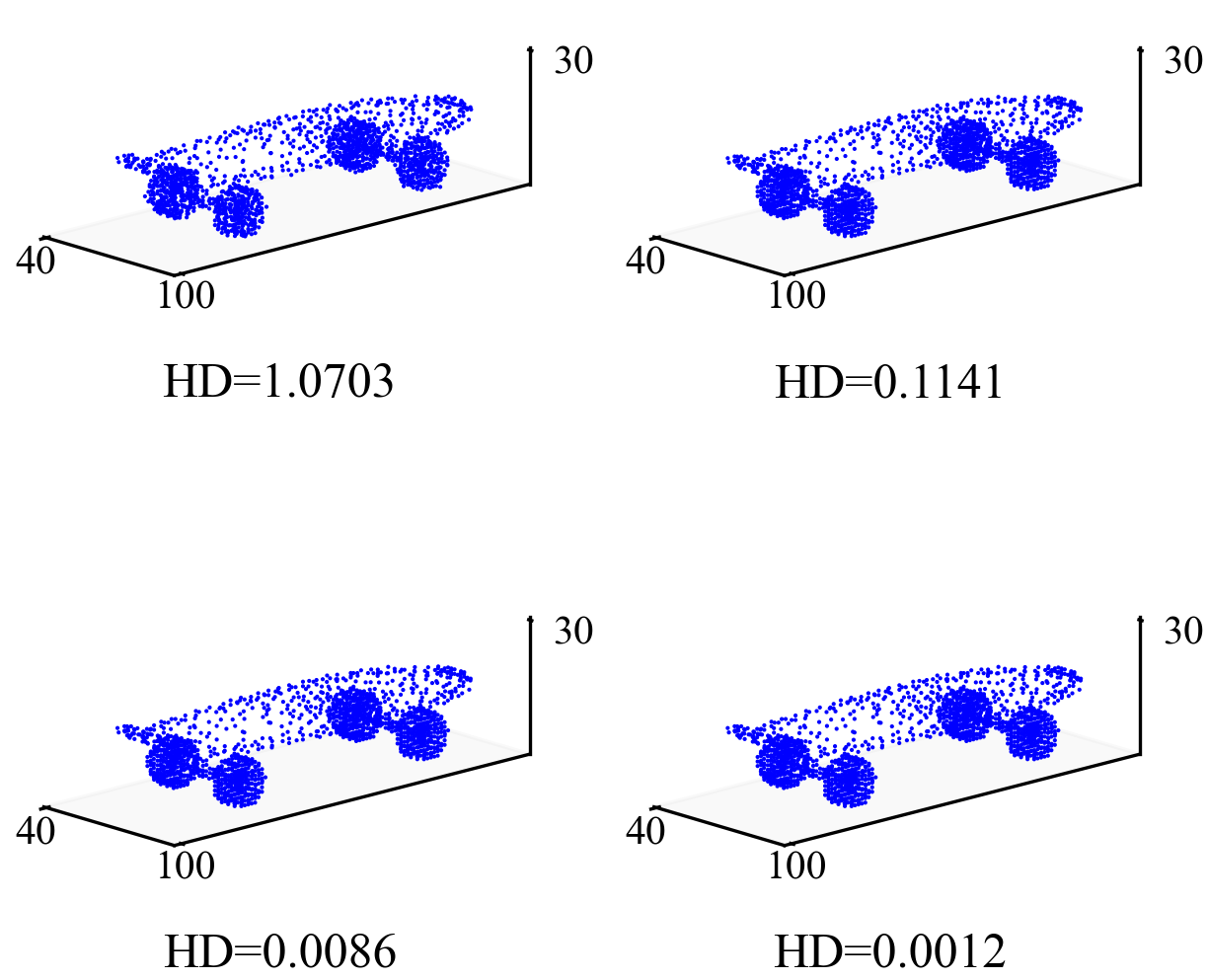}
        \caption{Skateboard, \url{https://youtu.be/a0fFu0z6BU0}}
    \end{subfigure}
    \caption{No FLS movement is visible with HD$<$0.1.  Click the URLs for a video.}\label{fig:visibleDiff}
\end{figure*}

\begin{figure}
\centering
\includegraphics[width=\columnwidth]{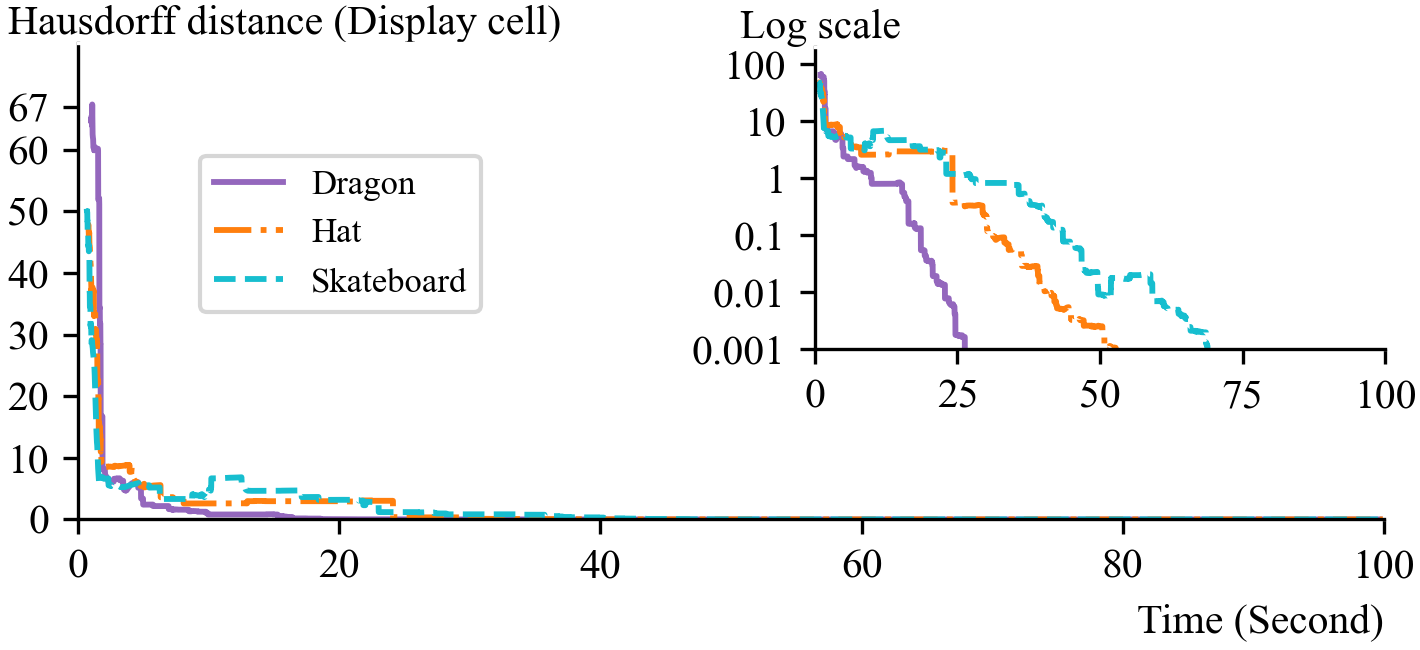}
\caption{HD with 3D shapes of Figure~\ref{fig:cmp3D}, $\epsilon$=5$\degree$.  The figure inside shows the HD in log scale.}
\label{fig:HDimpl}
\end{figure}

\begin{figure}

\begin{subfigure}[t]{\columnwidth}
\centering
\includegraphics[width=\textwidth]{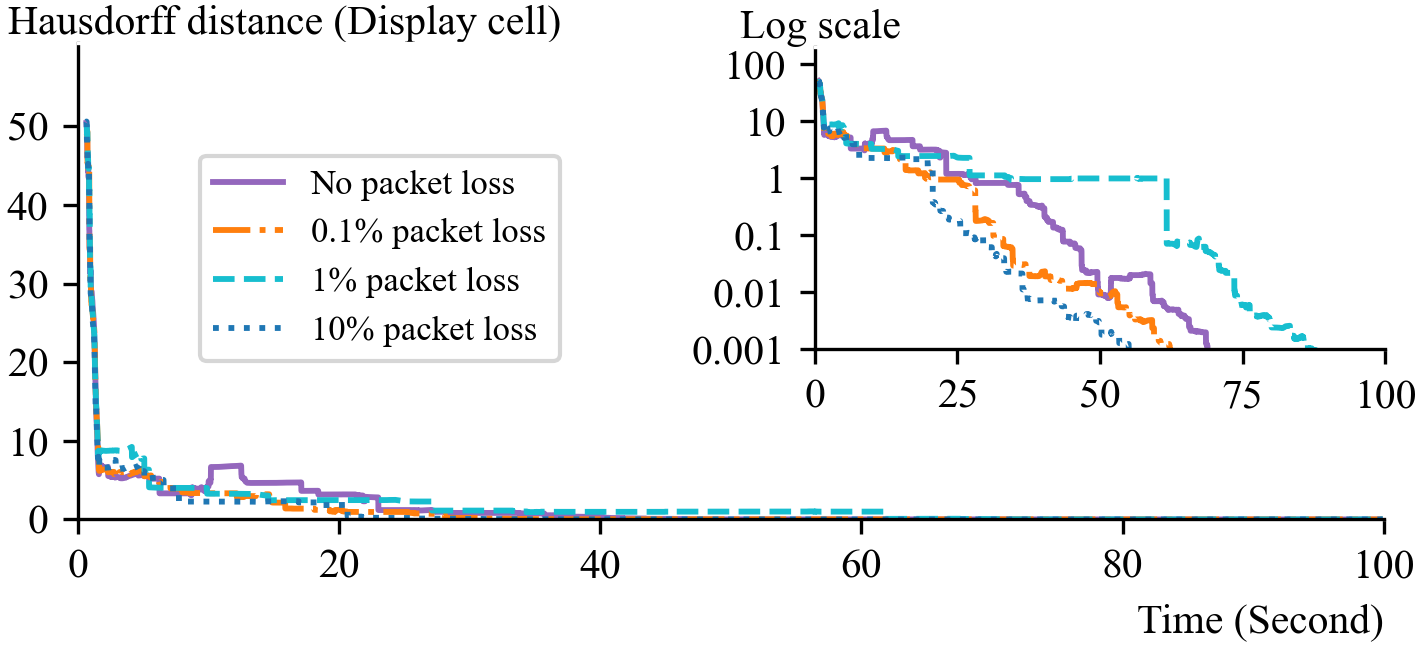}
\caption{Packet loss at receiver $P_{FLS}$.  The figure inside shows the HD in log scale.}
\label{fig:packetloss:percentages}
\end{subfigure}

\quad

\begin{subfigure}[t]{\columnwidth}
\centering
\includegraphics[width=\textwidth]{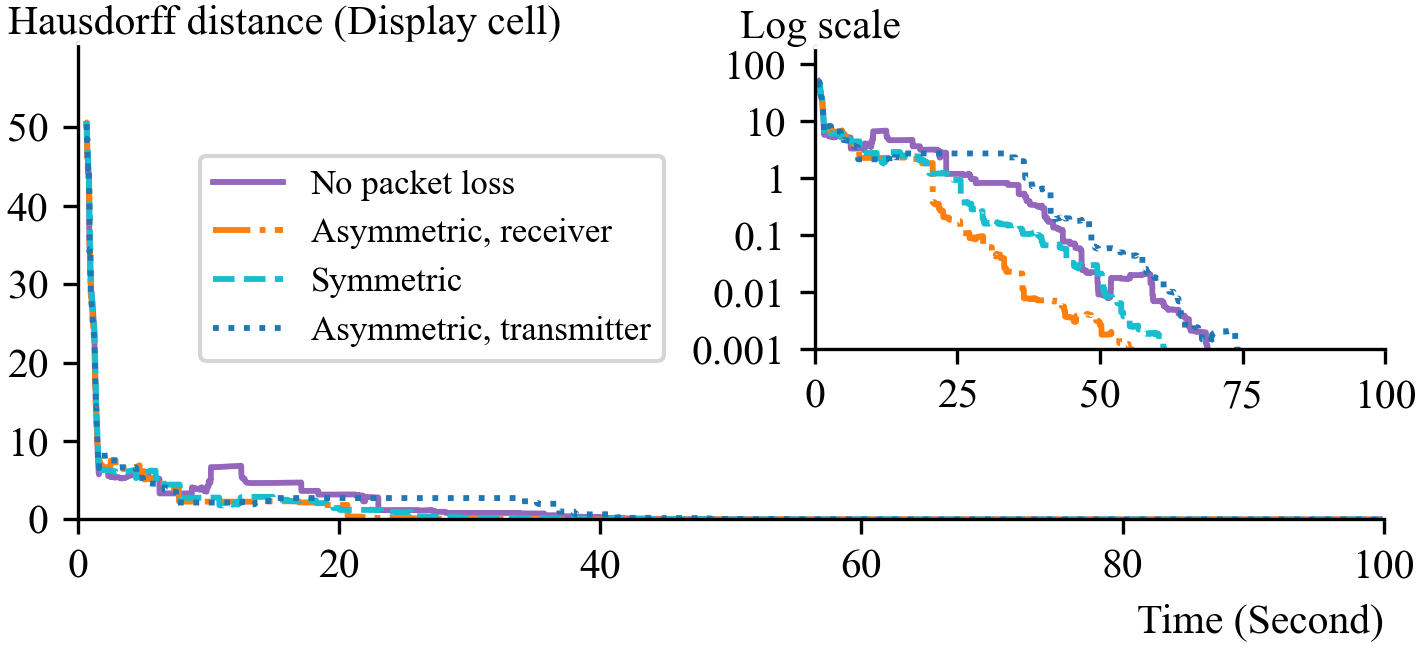}
\caption{10\% packet loss.  The figure inside shows the HD in log scale.}
\label{fig:packetloss:10}
\end{subfigure}

\caption{HD with packet loss, Skateboard, $\epsilon$=5$\degree$.}
\label{fig:packetloss}
\end{figure}

\noindent{\bf Obtained Results.}
Figure~\ref{fig:HDimpl} shows the HD as a function of time for the different shapes in Figure~\ref{fig:cmp3D}.
The y-axis of this figure is linear scale.
It reports HD as a function of the length of a display cell, a cube with a 5 Centimeter length.
Hence, a HD of 1 means the maximum distance between the points in the ground truth and the estimated truth is 5 Centimeter.
HD drops significantly in the first few seconds of SwarMer's execution because FLSs localize relative to one another.
When this distance drops below 50 Micrometer (0.005 Centimeter corresponding to HD of 0.001), we consider it close to zero and stop reporting the distance.
The different shapes arrive at this threshold at different times.  
See the variant of the Figure inside its empty space with y-axis in log scale.
The Dragon arrives at this threshold soonest as it consists of 760 FLSs.
The Skateboard requires the longest as it consists of more than twice (1727) as many FLSs.

Figure~\ref{fig:visibleDiff} shows the different shapes with HD values of 1, 0.1, and two HD values below 0.1.
There is no noticeable difference between the shapes with HD values lower than 0.1, i.e., a 5 millimeter error.

We designed the UDP-Wrapper to support both symmetric and asymmetric packet loss.  
It may be configured to drop packets with a pre-specified probability either at the transmitting, receiving, or both.
Figure~\ref{fig:packetloss} shows SwarMer is resilient to all kinds of network errors for the Skateboard. 
Depending on which event is lost, SwarMer may require a longer time to construct one swarm.
However, it continues to decrease the HD dramatically in the first few seconds of its execution.
We present the Skateboard because its reported HD for different forms of packet loss and loss rates are farthest apart, see the log scale version of the graph in Figure~\ref{fig:packetloss}.
They are alot closer for the Dragon and the Hat.

\begin{figure}
\centering
\includegraphics[width=\columnwidth]{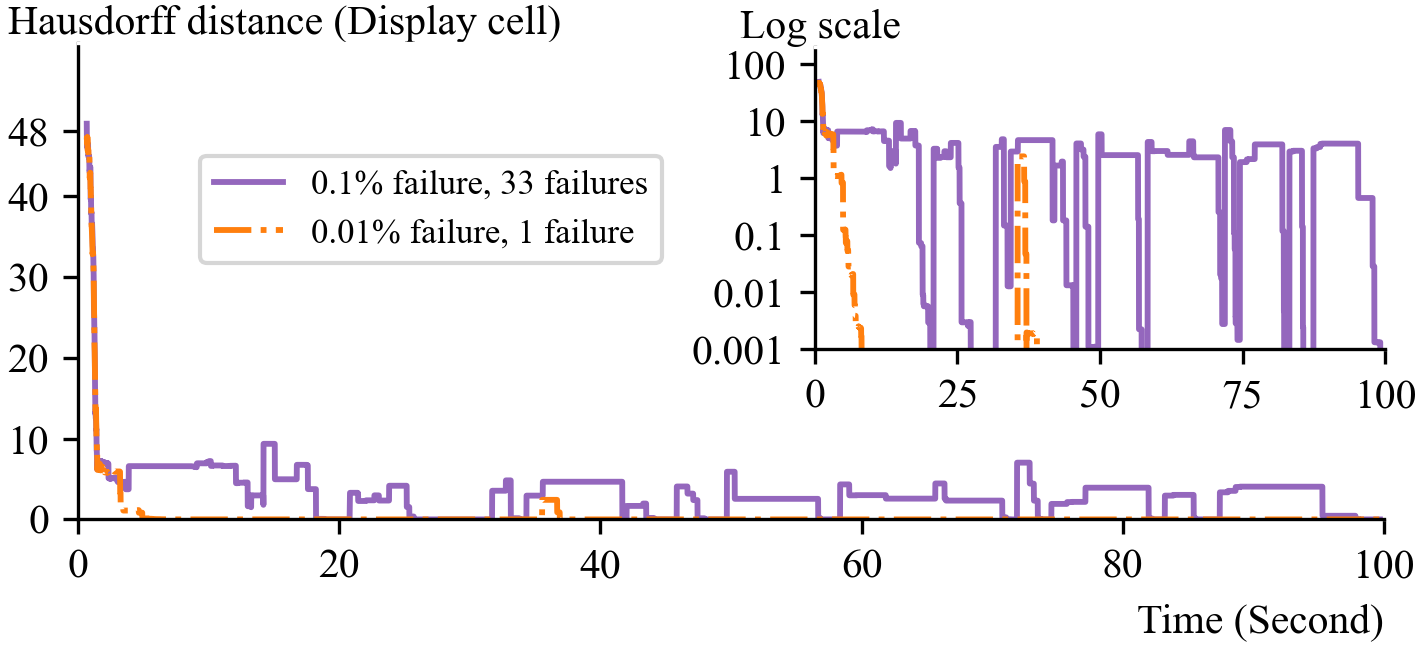}\hfill
\caption{HD with 0.1\% and 0.01\% failures per FLS per second for a 3D figure sampled with 200 points from the Chess piece.}
\label{fig:FLSfailures}
\end{figure}

\begin{figure}
\centering
\includegraphics[width=\columnwidth]{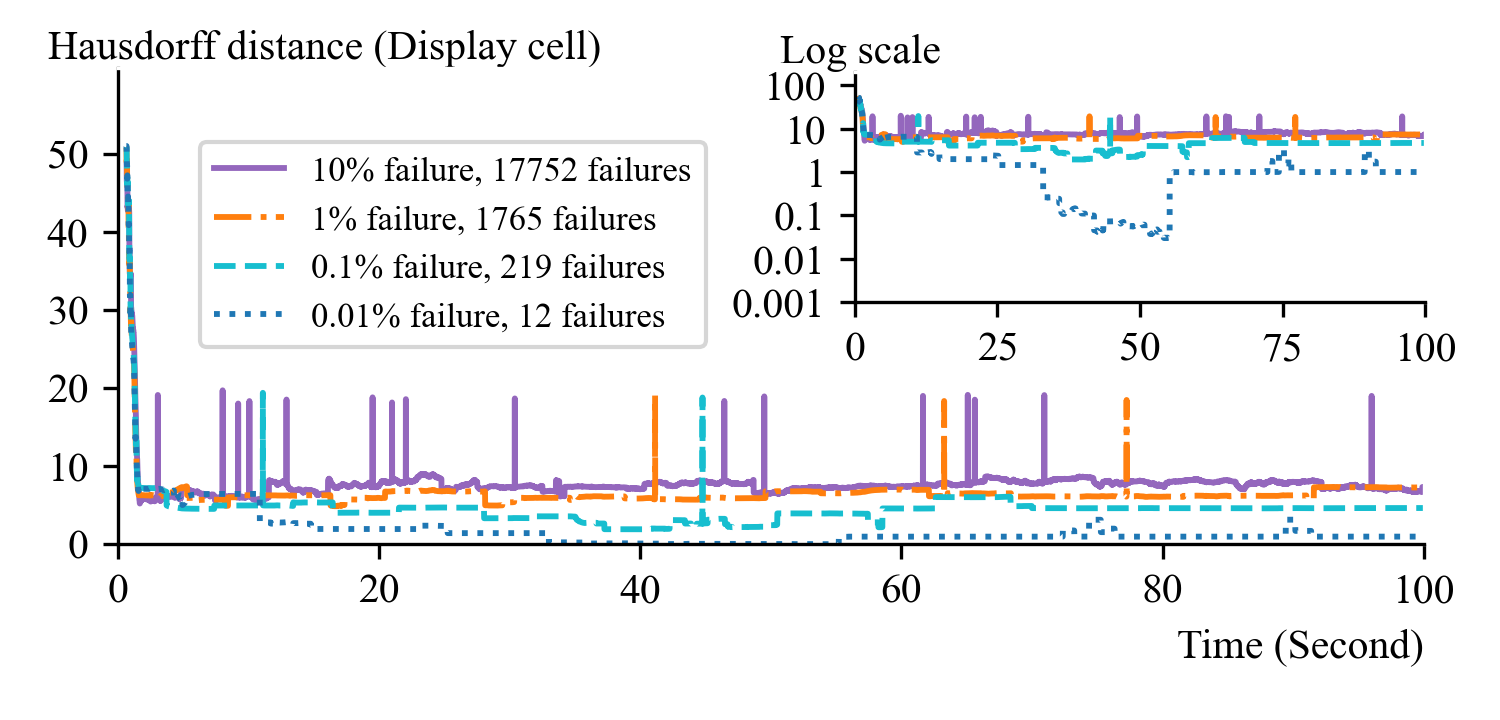}\hfill
\caption{HD with 0.01\%, 0.1\%, 1\%, and 10\% failures per FLS per second for the Skateboard.
Video clips available at 
\url{https://youtu.be/YE-hzpfONwg},
\url{https://youtu.be/1Tx-_DDZf0w},
\url{https://youtu.be/QEKJgNn0Yy8},
\url{https://youtu.be/mU-oIVaNu8M}, respectively.
}
\label{fig:FLSfailuresSK}
\end{figure}

SwarMer's use of leases enables it to tolerate FLS failures.
A replacement FLS deployed by a dispatcher may execute SwarMer to incorporate itself.  Figure~\ref{fig:FLSfailures} shows the HD with two different probabilities of failure per second per FLS:  0.1\% and 0.01\%.
The HD is different with 0.1\% and 0.01\% failures per second per FLS because there are a total of 33 and 1 failures during a 100 second execution of SwarMer.
The failure of an FLS causes a point to go dark. 
A dispatcher deploys its replacement FLS.
Once this FLS arrives at its coordinates using dead reckoning, it illuminates its assigned point.
This causes the HD to increase.
As the FLS localizes and merges into the swarm, the HD decreases.
This is shown as the sharp drop with the 0.01\$ failure.
There are more failures with 0.1\%.
The swarm is thawed when either all FLSs form one swarm or a fixed amount of time elapses.
This time is log (base 2) of the number of points, approximately 8 seconds in Figure~\ref{fig:FLSfailures}.
When a replacement FLS arrives at its coordinates, it may have to localize itself several times to reduce the HD. 
Each time, this FLS must wait for the swarm to thaw.
This explains the plateaus in the order of seconds in Figure~\ref{fig:FLSfailures}.

SwarMer is not a failure handling technique.
With a high failure rate, the resulting HD may be high.
Figure~\ref{fig:FLSfailuresSK} shows the HD with the Skateboard and failure rates\footnote{A failure of 10\% per FLS per second translates into one failure every 10 seconds.  In general, a failure of x\% per FLS per second translates into one failure every $\frac{100}{x}$ seconds.} as high as 10\%.
The HD is high with a failure rate of 1\% and higher because SwarMer incorporates many replacement FLSs concurrently.
The respective video clips show a distorted image.  
The display requires standby FLSs~\cite{shahram2022} to maintain a high quality of illumination.

\section{Related Work}\label{sec:related}
The SwarMer framework to localize FLSs is novel and to the best of our knowledge has not been described elsewhere.

Outdoor drone shows use GPS~\cite{gps2008} to localize and illuminate shapes in the sky.
The largest show consisted of 5200 drones to celebrate the 100th anniversary of the Communist Party of China in the night sky of Longgang, Shenzhen.
An FLS display is different because it does not have line of sight with GPS satellites.

The concept of a 3D display using FLSs to render multimedia shapes~\cite{shahram2021,shahram2022,mmsys2023,imeta2023,dv2023} is relatively new.
Prior studies either identify localization as a challenge~\cite{shahram2021,dv2023,imeta2023} or assume one without describing it~\cite{shahram2022,mmsys2023}. 
SwarMer is novel and designed for use by these studies.


Centralized indoor techniques such as optical motion capture systems merge images from fixed cameras positioned around a display volume to localize quadrotors and drones~\cite{opticalpositioning1,opticalpositioning2,preiss2017}.
These systems, e.g., Vicon, are highly accurate.
They require a unique marker arrangement for each drone and have enabled control and navigation of swarms of tens of drones.
It may be difficult (if not impossible) to form unique markers for more than tens of small drones measuring tens of millimeters diagonally~\cite{preiss2017}.
Hence their scalability is limited.
Moreover, these systems require broadcast at high frequency from a central computer to each drone~\cite{weinstein2018}.
This centralized communication channel is a single point of system failure.
It has a latency of several (7) milliseconds~\cite{preiss2017} and its bandwidth constrains the robustness and swarm size.
SwarMer is different because it is decentralized, requiring each FLS to localize relative to its neighbor to form a swarm.
The broadcast range of an FLS is controlled by its signal strength to communicate with a fixed number ($\eta$) of neighboring FLSs.

There exists a large number of studies in the area of formation control~\cite{opt2016,bearing-only2019,distFormation2020,angle2021}, use of sensors to localize robots and drones~\cite{uwb2015,uwb2016,uwbcao2021,
snaploc2019,vision2007,directodometry2018,markerless2010,BABINEC20141,xu2021,xu2022,wision2023,
md-track2018,orb2017,vins2018,openvins2020,lidar2017,loam2014,floam2021,loamlivox2020,viunet2023,uwbIMUvisual2021,liro2021}, and collision avoidance~\cite{clearpath2009,collisionfree2012,collisionfree2015,collisionavoidance2018,ReactiveCollisionAvoidance2008,ReactiveCollisionAvoidance2011,ReactiveCollisionAvoidance20112,reactiveColAvoidance2013,downwash1,downwash3,dcad2019,Engelhardt2016FlatnessbasedCF,navigation2017,reactiveColAvMorgan,reactiveColAvBaca,reactiveColAvMorganJ,speedAdjust2021,gameCollisionAvoidance2020,gameCollisionAvoidance2017,preiss2017,Ferrera2018Decentralized3C,planning2019,preiss2017whitewash,opticalpositioning1,navigate2014}.
These studies complement SwarMer and may be used by a system that implements a 3D display~\cite{dv2023}.
These prior studies lack the concept of FLSs forming a group (swarm), requiring a swarm to move together, swarms merging to become one, FLSs thawing their membership to repeat the process continuously, or use leases to explicitly support FLSs failing and leaving to charge their batteries.
These concepts, their decentralized design and implementation are novel and a contribution of SwarMer. 

\section{Conclusions and Future Research}\label{sec:future}
Inspired by swarms in nature, SwarMer is a decentralized technique for relative localization of FLSs to render complex 2D and 3D shapes.
It requires FLSs to localize relative to one another to form a swarm.
Movement of an FLS along a vector causes its entire swarm to move along the vector.
SwarMer uses the unique identifier of an FLS assigned at its deployment time to implement an organizational framework.
This framework compensates for the simplicity of individual FLSs to render complex 2D and 3D shapes accurately.

We presented a MATLAB simulation and a Python implementation of SwarMer.
The implementation shows SwarMer is fast.  
Within a few seconds, it reduces the HD to a value below 1 display cell.
Its continued execution reduces the HD further to match the ground truth more accurately.
It tolerates network packet loss, FLSs failing and leaving to charge their battery, and FLSs returning to illuminate points.

We are initiating an implementation of SwarMer by evaluating the relevant plugins to the SwarMer.
These include an investigation of the formation control techniques, alternative sensors for a localizing FLS to orient itself relative to an anchor FLS, and a collision avoidance technique as an FLS moves along a vector, see Section~\ref{sec:related}.

\section{Acknowledgments}
We thank Jiadong Bai and Shuqin Zhu for their valuable comments on the earlier drafts of this paper.
This research was supported in part by the NSF grant IIS-2232382.  We gratefully acknowledge CloudBank~\cite{cloudbank2021} and CloudLab~\cite{emulab} for the use of their resources to enable all experimental results presented in this paper.

\balance

\bibliographystyle{ACM-Reference-Format}
\bibliography{refs}  

\end{document}